\documentclass[letterpaper]{article} 
\usepackage{aaai23}  
\usepackage{times}  
\usepackage{helvet}  
\usepackage{courier}  
\usepackage[hyphens]{url}  
\usepackage{graphicx} 
\usepackage{natbib}  
\usepackage{caption} 
\graphicspath{{./}{figures/}{figures/mnist/}}

\usepackage{etoolbox}
\newtoggle{arxiv}
\settoggle{arxiv}{true}  

\title{Generalizing Downsampling from Regular Data to Graphs%
\thanks{\iftoggle{arxiv}{}{Supplementary Material can be found at \protect\url{https://arxiv.org/abs/2208.03523}.} We would\xspace\iftoggle{arxiv}{}{also} like to thank Federico Poloni and Federico Errica for their most useful suggestions on earlier versions of this paper. This research was partially supported by TAILOR, a project funded by EU Horizon 2020 research and innovation
programme under GA No 952215.}%
}

\author{
	Davide Bacciu, Alessio Conte, Francesco Landolfi
}
\affiliations{
	Department of Computer Science, Università di Pisa
}

\usepackage{amsmath,amsfonts,amsthm,amsbsy,amssymb}
\usepackage{graphicx}
\usepackage{textcomp}
\usepackage{xcolor}
\usepackage{etoolbox}

\usepackage[utf8]{inputenc} 
\usepackage{url}            
\usepackage{booktabs}       
\usepackage{amsfonts}       
\usepackage{nicefrac}       
\usepackage{microtype}      
\usepackage{xcolor}         
\usepackage{xspace}
\usepackage{dsfont}
\usepackage{pifont}

\usepackage{graphicx}
\usepackage{caption}
\usepackage{subcaption}
\usepackage{bm}
\usepackage{algorithm}
\usepackage[noend]{algpseudocode}
\usepackage[inline]{enumitem}

\setlist[itemize]{leftmargin=*}
\setlist[enumerate]{leftmargin=*}

\usepackage{pgfplots}
\pgfplotsset{compat=1.16}

\algblock{ParFor}{EndParFor}
\algnewcommand\algorithmicparfor{\textbf{for}}
\algnewcommand\algorithmicpardo{\textbf{do in parallel}}
\algnewcommand\algorithmicendparfor{\extbf{end for}}
\algrenewtext{ParFor}[1]{\algorithmicparfor\ #1\ \algorithmicpardo}
\algrenewtext{EndParFor}{\algorithmicendparfor}

\makeatletter
\ifthenelse{\equal{\ALG@noend}{t}}%
  {\algtext*{EndParFor}}
  {}%
\makeatother

\makeatletter
\renewcommand{\ALG@beginalgorithmic}{\footnotesize}
\makeatother

\makeatletter
\patchcmd{\maketag@@@}{\normalfont}{\normalfont\footnotesize}{}{} 
\makeatother

\usepackage[capitalize,nameinlink]{cleveref}

\newtheorem{theorem}{Theorem}
\newtheorem{lemma}{Lemma}
\newtheorem{corollary}{Corollary}
\newtheorem{proposition}{Proposition}

\newtheorem{remark}{Remark}
\newtheorem*{remark*}{Remark}
\newtheorem{definition}{Definition}

\graphicspath{{figures/}}

\newlist{thmlist}{enumerate}{1}
\setlist[thmlist]{
    label=\textit{\bfseries (\alph{thmlisti})}, 
    ref=\thetheorem\alph{thmlisti},
    itemsep=0pt,
    topsep=0pt
}

\newlist{thmlist*}{enumerate*}{1}
\setlist[thmlist*]{
    label=\textit{\bfseries (\alph{thmlist*i})}, 
    ref=\thetheorem\alph{thmlist*i}
}

\Crefname{thmlisti}{Theorem}{Theorems}

\newcommand{\nat}{\ensuremath\mathds{N}}
\newcommand{\real}{\ensuremath\mathds{R}}
\newcommand{\mat}[1]{\ensuremath\boldsymbol{\mathbf{#1}}}
\newcommand{\tr}{\ensuremath{\!\mat{\top}\!}}
\renewcommand{\vec}[1]{\ensuremath\boldsymbol{\mathbf{#1}}}
\newcommand{\gr}[1]{\ensuremath #1}
\newcommand{\degree}{\ensuremath \operatorname{deg}}
\newcommand{\G}{\gr{G}}
\renewcommand{\P}{\ensuremath\gr{P}}
\renewcommand{\H}{\gr{H}}
\newcommand{\neigh}{\ensuremath N} 
\newcommand{\rneigh}{\ensuremath\widehat{\neigh}}
\newcommand{\adj}{\ensuremath{\mat{A}}}
\newcommand{\Smat}{\ensuremath\mat{S}}

\newcommand{\length}{\ensuremath\operatorname{d}}

\newcommand{\rank}{\ensuremath{\pi}}
\newcommand{\bigO}[1]{\ensuremath{{O(#1)}}}
\newcommand{\reduce}{\ensuremath\mathcal{R}}
\newcommand{\I}{\ensuremath\mat{I}}
\newcommand{\model}[1]{\textsc{#1}\xspace}
\newcommand{\repr}{\ensuremath\rho}

\DeclareMathAlphabet{\mathbfsf}{\encodingdefault}{\sfdefault}{bx}{n}
\newcommand{\tens}[1]{\ensuremath\mathbfsf{#1}}
\newcommand{\fullgrid}{\ensuremath\operatorname{G}_{\boxtimes}}

\newcommand{\topk}[1][k]{{top}-${#1}$\xspace}

\newcommand{\kmis}[1][k]{${#1}$-MIS\xspace}
\newcommand{\asapool}{\model{ASAPool}}

\begin{document}

\maketitle

\begin{abstract}
Downsampling 
produces coarsened, multi-resolution representations of data and it is used, for example, to produce lossy compression and visualization of large images, reduce 
computational costs, and boost deep neural representation learning.
Unfortunately, due to their lack of a regular structure, there is still no consensus on how downsampling should apply to graphs and linked data. 
Indeed reductions in graph data are still needed for the goals described above, but reduction mechanisms do not have the same focus on preserving topological structures and properties, while allowing for resolution-tuning, as is the case in regular data downsampling.

In this paper, we take a step in this direction, introducing a unifying interpretation of downsampling in regular and graph data. In particular, we define a graph coarsening mechanism which is a graph-structured counterpart of controllable equispaced coarsening mechanisms in regular data. We prove theoretical guarantees for distortion bounds on path lengths, as well as the ability to preserve key topological properties in the coarsened graphs. We leverage these concepts to define a graph pooling mechanism that we empirically assess in graph classification tasks, providing a greedy algorithm that allows efficient parallel implementation on GPUs, and showing that it compares favorably against pooling methods in literature. 

\end{abstract}

\section{Introduction}

The concept of information coarsening is fundamental in the adaptive processing of data, as it provides a simple, yet effective, means to obtain multi-resolution representations of information at different levels of abstraction. In large scale problems coarsening also serves to provide computational speed-ups by solving tasks on the reduced representation, ideally with a contained loss in precision with respect to solving the original problem. 

Coarsening is key in Convolutional Neural Networks~\cite[CNNs,][]{fukushima_neocognitron_1980,lecun_backpropagation_1989}, where pooling is often used to repeatedly subsample an image to extract visual feature detectors at increasing levels of abstraction (e.g., blobs, edges, parts, objects, etc). Downsampling is also popular in the adaptive processing of timeseries where, for instance, it is used in clockwork-type Recurrent Neural Networks~\cite{koutnik_clockwork_2014,carta_incremental_2021} to store information extracted at different frequencies and timescales. More recently, the Graph Convolutional Networks~\cite[GCNs,][]{micheli_neural_2009,gori_new_2005,bacciu_gentle_2020} community popularized graph reduction mechanisms as a structured counterpart of the image pooling mechanism in \emph{classical} CNNs.  

The definition of a reduction mechanism that downsamples information 
at regular intervals between data points (e.g., a sample, a pixel, a timestamped observation, etc) 
is straightforward when working with images and time series. It can be achieved simply by picking up a data point every $k$ ones, where $k$ is a given reduction factor defining the distance between the sampled points in the original data, possibly aggregating the properties of non-selected point with appropriate functions. The same approach cannot be straightforwardly applied to graphs, which lack regularity and a consistent ordering among their constituent data points, i.e., the nodes. Therefore, defining a well-formed notion of downsampling for graphs becomes non-trivial. 
The research community has been tackling this issue by a number of approaches, including differentiable clustering of node embeddings~\cite{ying_hierarchical_2018,bianchi_spectral_2020}, 
graph reductions~\cite{shuman_multiscale_2016,loukas_graph_2019}, and 
node ranking~\cite{cangea_towards_2018,gao_graph_2019}.
Notably, approaches like the latter select important nodes in a graph and simply discard the rest without protecting the linked structure of the network,
while reduction methods typically focus on preserving structure without accounting for the role or relevance of nodes involved.

What is yet an open problem is how to define a controllable graph coarsening method, which reduces the size while preserving the overall structure by sampling \textit{representative} yet \emph{evenly} spaced elements, similarly to the approaches discussed above for image and time series reduction.
%

This paper provides a first approach introducing such a topology-preserving graph coarsening and its use in graph pooling. We provide mechanisms which are the graph equivalent of pooling and striding operators on regular data, accompanying our intuition with formal proofs (in the Supplementary Material) of the equivalence of such operators on graphs which model regular data. 

Central to our contribution is the definition of a mechanism to find a set of nodes that are approximately equally spaced (at distance no less than $k$) in the original graph. We build on the graph-theoretic concept of Maximal $k$-Independent Sets (\kmis), that also comes with the ability to pin-point important nodes in each area of the graph. The selected nodes are then used as vertices of the reduced graph whose topology is defined in such a way that key structural properties of the original graph are well preserved. To this end, we provide theoretical guarantees regarding distance distortions between a graph and its reduction. 
Additionally, we prove the reduced graph has the same number of connected components as the original.
The latter point is particularly relevant for a graph pooling mechanism as it guarantees that the structure is not broken in disconnected fragments, which can hinder the performance of neural message passing in the GCN layers. 

Such properties are fundamental to ensure that the original graph is downsampled evenly throughout its structure, preserving distances and sparseness of the key focal points in the graph. By this means, the reduced graph can be used as an accurate fast estimator of the distances between nodes in the original graph, where the amount of compression 
can be easily regulated through the choice of the $k$ reduction factor. 

Concurrently, we borrow from node-ranking methods~\cite{gao_graph_2019} to produce {\kmis}s that maximize the total weights associated to the selected nodes, 
in order to preserve relevant nodes without compromising structure. 


In summary, our contributions 
are the following:
\begin{itemize}
    \item We introduce a graph coarsening method leveraging \kmis that is the graph-structured counterpart of equispaced sampling in flat data. 
    We provide a greedy parallel algorithm to efficiently compute the \kmis reduction, which is well suited to use in GPU accelerators
    (\cref{sect:kmis}).
    \item We give formal proof of equivalence of our approach to regular downsampling in convolutional neural networks, when applied to diagonal grid graphs (\cref{sect:theo} and Supplementary Material).
    \item We prove theoretical guarantees on the distance distortions between a graph and its reduction. We provide also a formal complexity analysis of the introduced algorithms, proving, both theoretically and experimentally, their scalability on large real-world graphs (\cref{sect:theo} and Supplementary Material).
    \item We integrate \kmis reduction both as a pooling layer and as a downsampling operator for GCNs, providing an empirical confirmation of its advantages over literature approaches on graph classification benchmarks (\cref{sect:exp}).
\end{itemize}

\section{Notation and Definitions}\label{sect:back}

We represent a graph $\G$ as a pair of disjoint sets $(V, E)$, where $V = \{1, \dots, n\}$ 
is its node set and $E \subset V\times V$ its edge set, with $|E| = m$. 
%
%
%
A  graph can also be represented as a symmetric matrix $\adj \in \real_+^{n\times n}$, such that $\adj_{uv} = \adj_{vu}$ is equal to a weight associated to the edge $uv \in E$ or zero if $uv \not\in E$.
The neighborhood $\neigh(v)$ of $v$ is the set of nodes adjacent to it (denoted $\neigh[v]$ if includes $v$ itself), and the degree $\degree(v)$ of $v$ is defined as the number of its neighbors, i.e., $\degree(v) = |\neigh(v)|$. 
%
%
The \emph{unweighted} distance between two nodes $u, v \in V$, denoted as $\length(u, v)$, is defined as the length of the shortest path between the two nodes. If there is no path between the two nodes, then $\length(u,v) = 
\infty$. 
%
The $k$-hop neighborhood $\neigh_k(v)$ of $v$ ($\neigh_k[v]$ if inclusive) is the set of nodes that can be reached by a path in $\G$ of length at most $k$. The \emph{$k$-th power} of a graph $\G^k$ is the graph where each node of $\G$ is connected to its $k$-hop neighbors.
To avoid confusion, any function may be denoted with a subscript to specify the graph on which is defined (e.g., 
$\length_\G$).
An \emph{independent set}, is a set of nodes $S \subseteq V$ such that no two of which are adjacent in $\G$. An independent set is \emph{maximal} if 
is not a subset of another one in $\G$.
A (maximal) 
$k$-\emph{independent set} is a (maximal) 
independent set of $\G^k$.

\section{Graph Coarsening with {$\bm{k}$}-MWIS}\label{sect:kmis}

When dealing with signals, images, or other kinds of Euclidean data, \emph{downsampling} often amounts to \emph{keeping every $k$-th} data point, where $k$ is a given reduction factor. This means, for a generic discrete $n$-dimensional Euclidean datum, keeping a subset of its points such that every two of them are \emph{exactly} $k$ points far from each other on every of its dimensions. On graph-structured data, we lose this regularity along with the concept of dimensionality, and hence defining a new notion of downsampling that applies to graph becomes non-trivial. 

Here we define a graph coarsening method that, similarly to {\it classical} downsampling, reduces the size of a graph $\G$ by a given ``factor'', by finding a set of \emph{almost} evenly spaced nodes within $\G$. These nodes will form the node set of the reduced graph, while its topology will be constructed starting from $\G$ in a way in which some of its key properties will be preserved, such as connectivity, or approximated, such as pairwise node distances. 


\paragraph{Coarsening algorithm.}

Given a graph $\G = (V, E)$ and a distance $k$, we want to obtain a coarsen representation of $\G$ by first selecting a set of nodes $S \subseteq V$, that we refer to as \emph{centroids}, such that every two centroids are more than $k$ hops distant from each other, and such that no area of the graph remains unsampled; in other words, a \emph{maximal $k$-independent sets} (\kmis) of $\G$: this way, each centroid will be more than $k$ hops from every other, while the \emph{maximality} ensures every node of $G$ is within $k$ hops from a centroid.

%
%
Any MIS of a graph $\G^k$ is a \kmis of $\G$~\cite{agnarsson_powers_2003}, thus a \kmis could be na\"ively computed by known MIS algorithms, such as \citet{luby_simple_1985} or \citet{blelloch_greedy_2012}, on the $k$-th power of the adjacency matrix of $\G$. Using this approach will require $\bigO{n^2}$ space since the density of $\G^k$ increases rapidly with $k$, becoming rapidly impractical for real world graphs with millions or billions of nodes.
%
To overcome this problem, we introduce  \cref{alg:k-mis} that efficiently computes a \kmis of $\G$ without explicitly computing its $k$-th power.
\begin{table}[tb]
\setlength{\intextsep}{0pt}  
    \begin{algorithm}[H]
    \begin{algorithmic}[1]
    	\Function{\kmis}{$\G$, $U$, $\rank$}
    	\State\algorithmicif\ $\lvert U \rvert = 0$ \algorithmicthen\ \Return $\emptyset$\label{line:k-mis-check}
    	\State $\rank_0 \gets \rank$\label{line:k-hop-1}
    	
    	\For{$i = 1, \dots, k$}\label{line:k-hop-2}
    	    \ParFor{$v \in U$}\label{line:k-hop-2-1}
    	    \State $\rank_i(v) \gets \min_{u \in \neigh[v]\cap U}\ \rank_{i-1}(u)$\label{line:k-hop-2-2}
    	    \EndParFor
    	\EndFor
    	\State $S_0 \gets \{v \in U \mid \rank(v) = \rank_k(v)\}$\label{line:k-hop-3}
    	\For{$i = 1, \dots, k$}\label{line:k-restrict-2}
    	    \State $S_i \gets \bigcup_{v \in S_{i-1}} \neigh[v]$\label{line:k-restrict-2-1}
    	\EndFor
    	\State $R \gets U \setminus S_k$\label{line:k-restrict-3}
    	\State \Return $S_0 \cup {}$\Call{\kmis}{$G$, $R$, $\rank$}\label{line:k-mis-rec}
    	\EndFunction 
    \end{algorithmic}
    \caption{ Parallel Greedy $k$-\textsc{MIS} algorithm, adapted from~\citet{blelloch_greedy_2012}. Given a graph $\G$, a subset of its nodes $U \subseteq V$, and a node ranking $\rank$, returns a maximal $k$-independent set in $\G$, with $k \in \nat$.}
    \label{alg:k-mis}
    \end{algorithm}\vspace{1pt}
    \begin{algorithm}[H]
    \begin{algorithmic}[1]
    	\Function{Cluster}{$\G = (V, E)$, $k$, $\rank$}
    	\State $S \gets$ \Call{\kmis}{$\G$, $V$, $\rank$}
    	\State $\rank_0 \gets \rank$
    	
        \ParFor{$v \in V \setminus S$}
            \State $\rank_0(v) \gets +\infty$
        \EndParFor
    	
    	\For{$i = 1, \dots, k$}
    	    \ParFor{$v \in V$}
    	    \State $\rank_i(v) \gets \min_{u \in \neigh[v]}\ \rank_{i-1}(u)$
    	    \EndParFor
    	\EndFor
    	\State \Return $\{\{u \in V \mid \rank_k(u) = \rank(v)\}\}_{v \in S}$
    	\EndFunction 
    \end{algorithmic}
    \caption{Parallel \kmis partitioning algorithm. Given a graph $\G$, $k\in \nat$, and a node ranking $\rank$, returns a partition of $\G$.%
    }
    \label{alg:cluster}
    \end{algorithm}
\end{table}
Once the \kmis $S \subseteq V$ is computed with \cref{alg:k-mis}, we construct the coarsened graph $\H = (S, E')$ as follows:
\begin{enumerate}
    \item using \cref{alg:cluster}, we compute a partition $\mathcal{P}$ of $V$ of size $\lvert S \rvert$, such that
    \begin{enumerate}
        \item every $P\in\mathcal{P}$ contains exactly one centroid and (a subset of) its $k$-hop neighbors, and
        \item for every node in $P$ there is a centroid in $P$ at distance at most $k$-hops;
    \end{enumerate}
    
    \item\label{item:edges} for every edge in $E$ we add an edge in $E'$ joining the two nearest centroids in the partitions containing the source and destination nodes. If this generates multiple edges, we coalesce them into a single one, and we aggregate their weights according to a predefined aggregation function (e.g., sum);
    
    \item \emph{(pooling, optional)} in case of weights/labels associated to the nodes, these can also be aggregated according to the partitioning $\mathcal{P}$. 
\end{enumerate}
A detailed discussion of \cref{alg:k-mis,alg:cluster} will be provided later in \cref{sect:theo}.
\begin{figure}[tb]
    \centering
    \resizebox{\linewidth}{!}{%
    \def\imsize{0.3\linewidth}
    \centering
    \begin{tabular}{@{}r@{\hspace{1ex}}c@{}c@{}c@{}c@{}}
    & \begin{minipage}{\imsize}
            \centering \footnotesize
            \model{Pooling, Mean}\\
            $(p = k + 1)$
        \end{minipage}
    & \begin{minipage}{\imsize}
            \centering \footnotesize
            \model{Ours, Mean}\\
            (\emph{lexicographic})
        \end{minipage}
    & \begin{minipage}{\imsize}
            \centering \footnotesize
            \model{Ours, Mean}\\
            (\emph{intensity})
        \end{minipage}
    & \begin{minipage}{\imsize}
            \centering \footnotesize
            \model{Ours, Strided}\\
            (\emph{intensity})
        \end{minipage}\\[1.5ex]
    \rotatebox{90}{\small\quad\quad\ \ \ $k = 0$}
    &\includegraphics[width=\imsize]{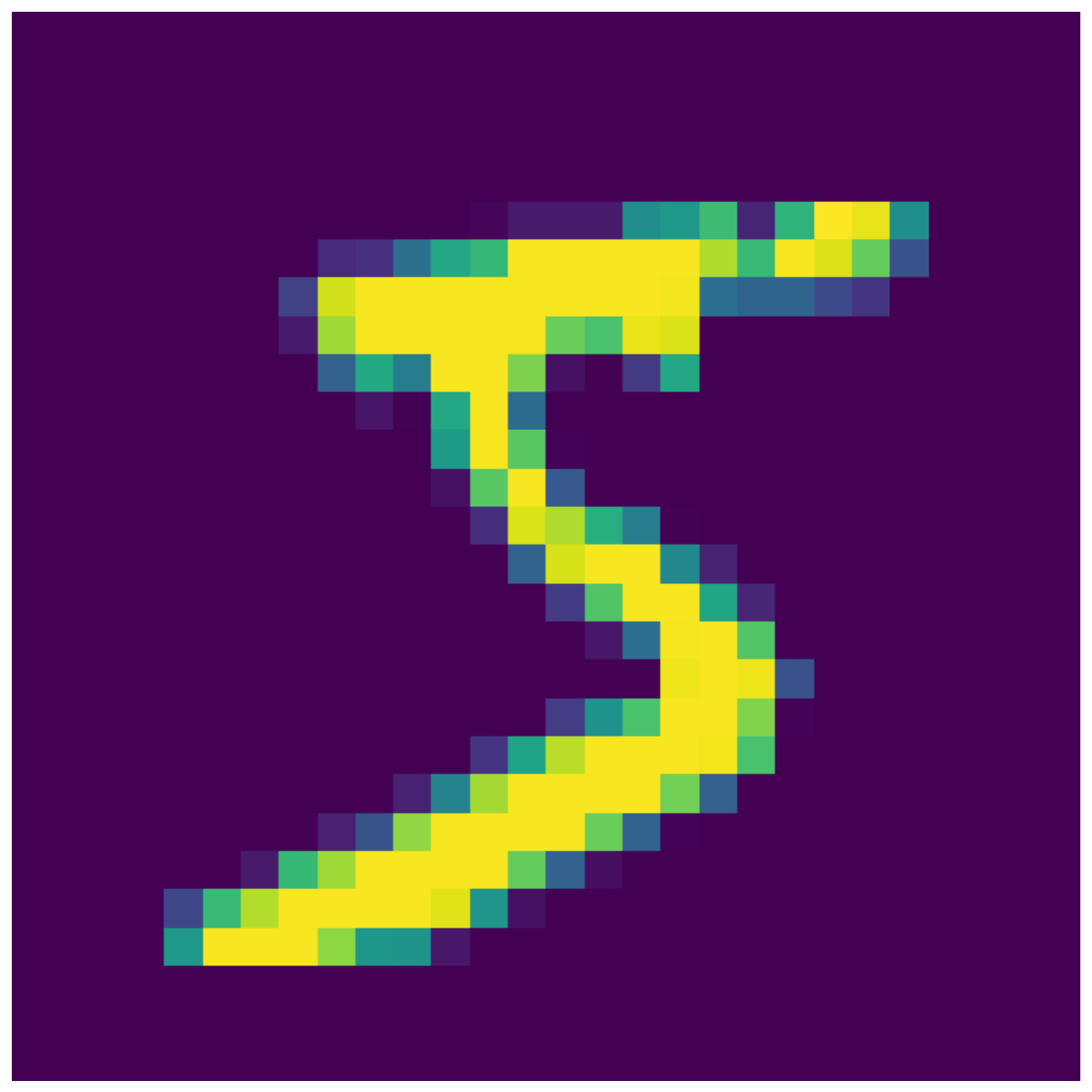}
    &\includegraphics[width=\imsize]{mnist/mnist_POOL1.pdf}
    &\includegraphics[width=\imsize]{mnist/mnist_POOL1.pdf}
    &\includegraphics[width=\imsize]{mnist/mnist_POOL1.pdf}\\[-4pt]
    \rotatebox{90}{\small\quad\quad\ \ \ $k = 1$}
    &\includegraphics[width=\imsize]{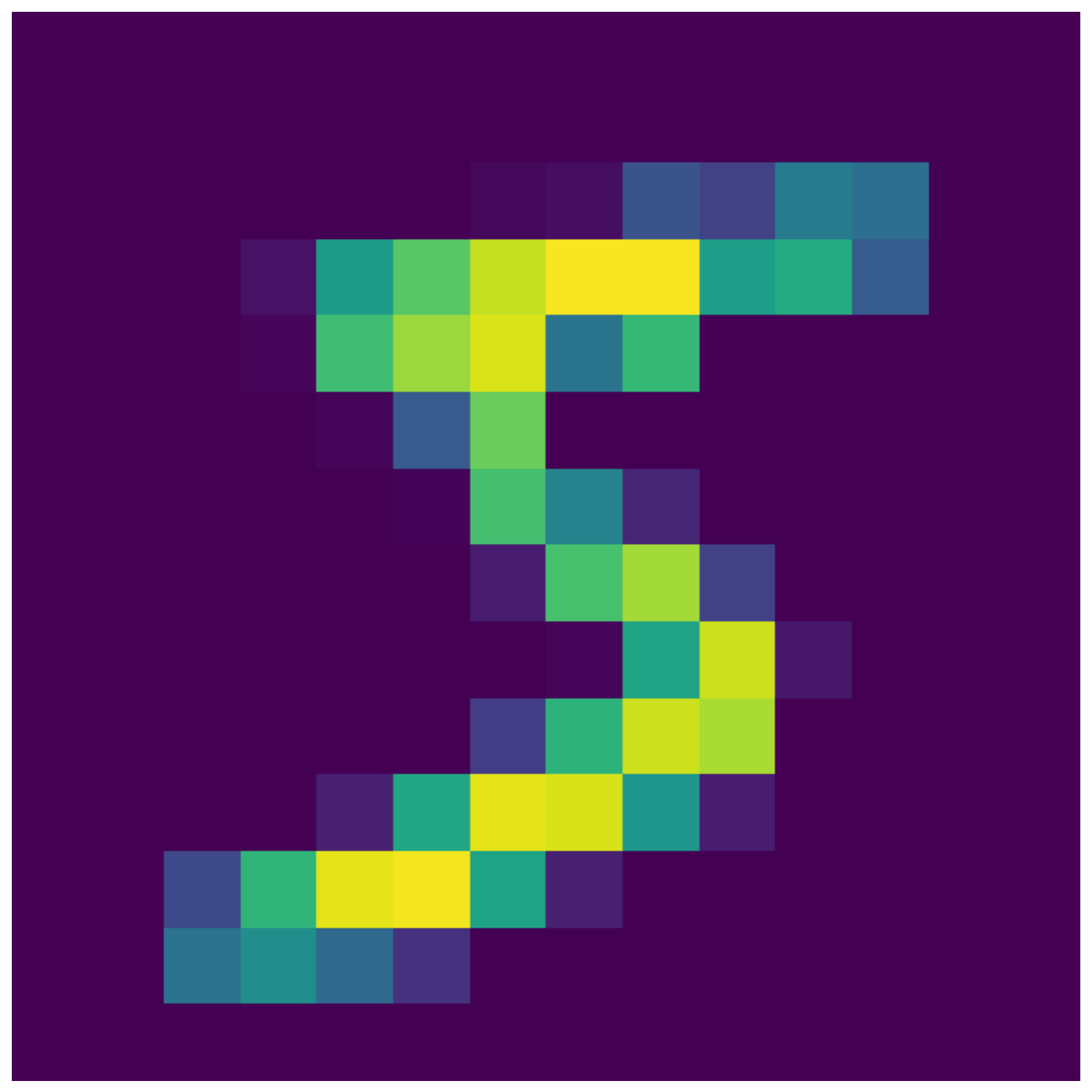}
    &\includegraphics[width=\imsize]{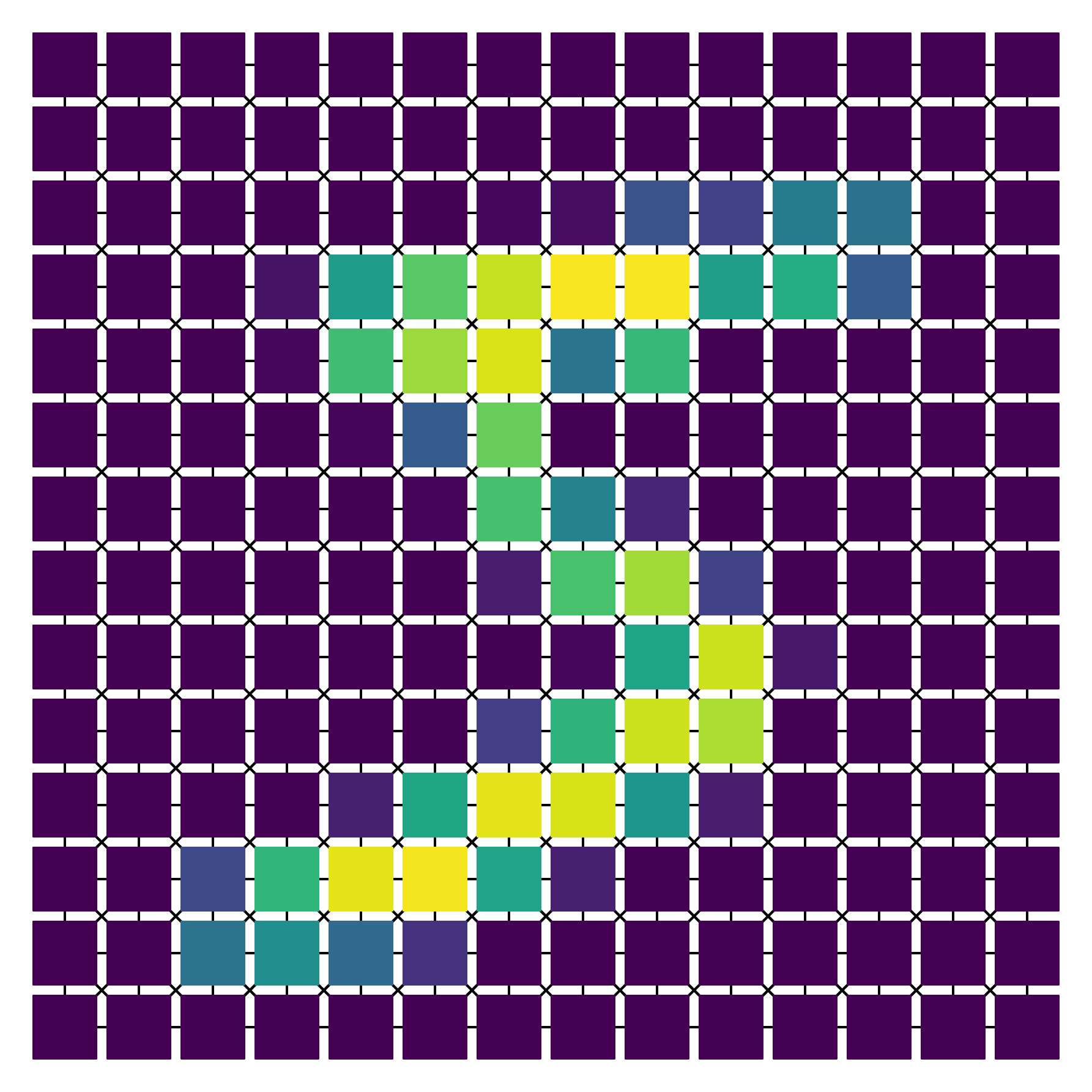}
    &\includegraphics[width=\imsize]{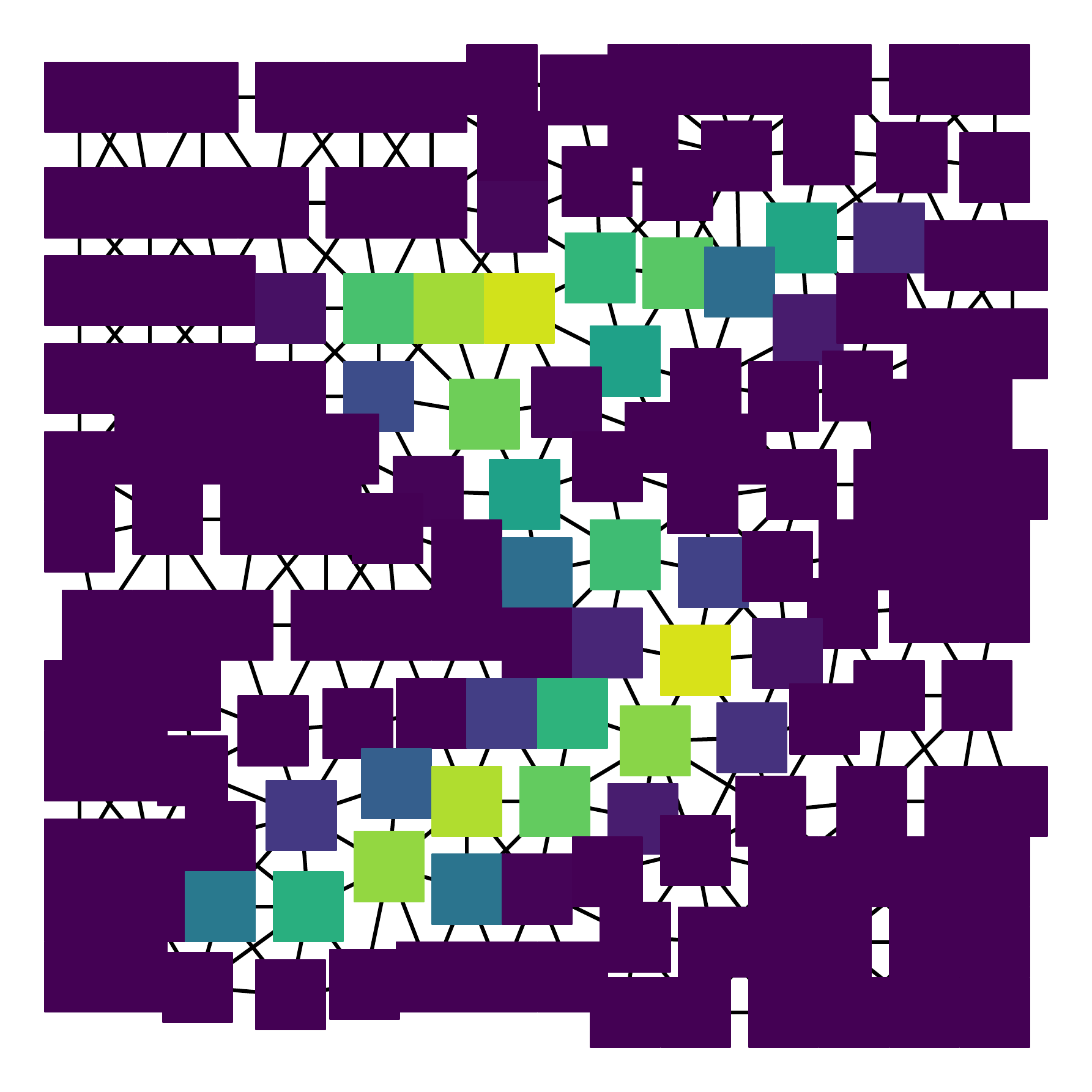}
    &\includegraphics[width=\imsize]{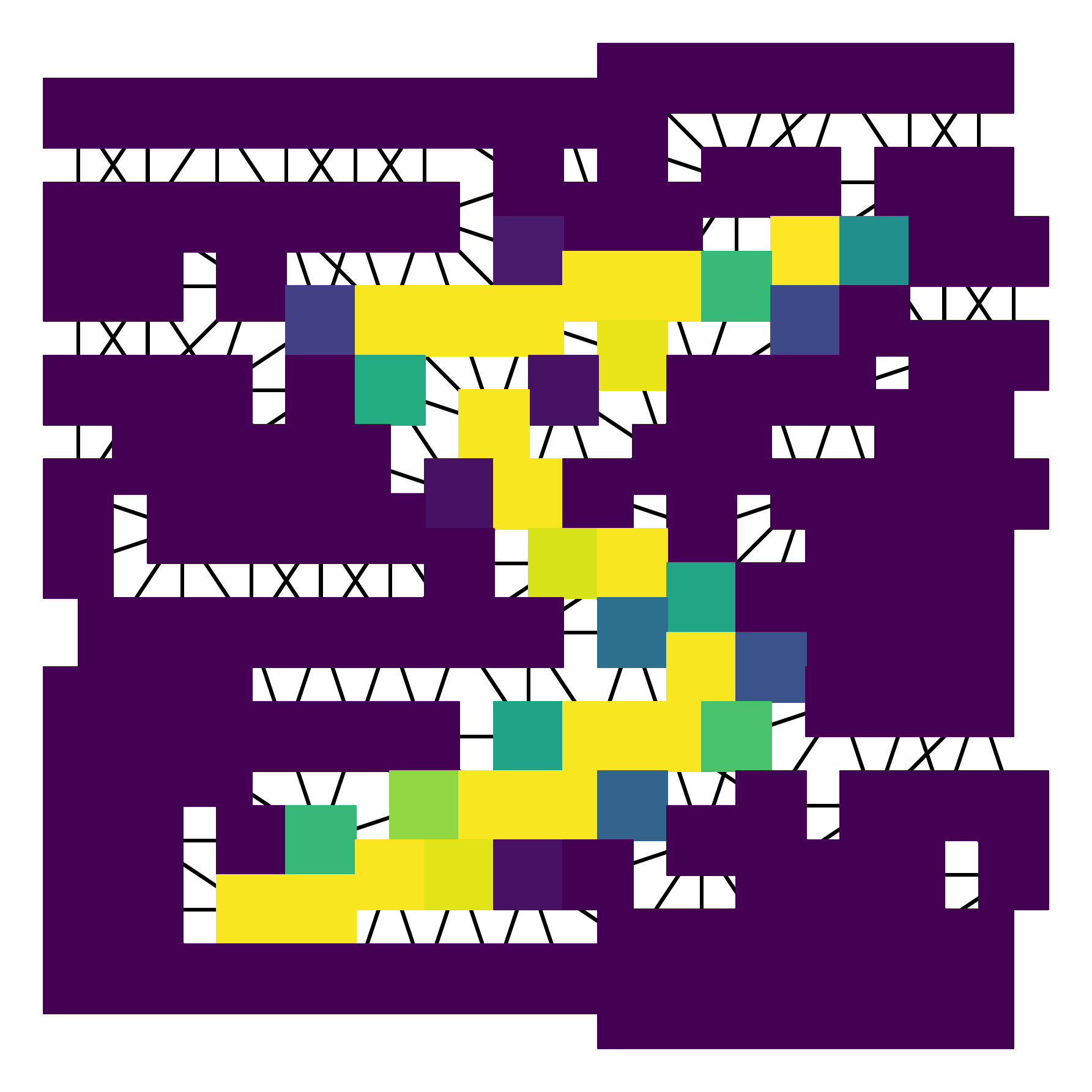}\\[-4pt]
    \rotatebox{90}{\small\quad\quad\ \ \ $k = 2$}
    &\includegraphics[width=\imsize]{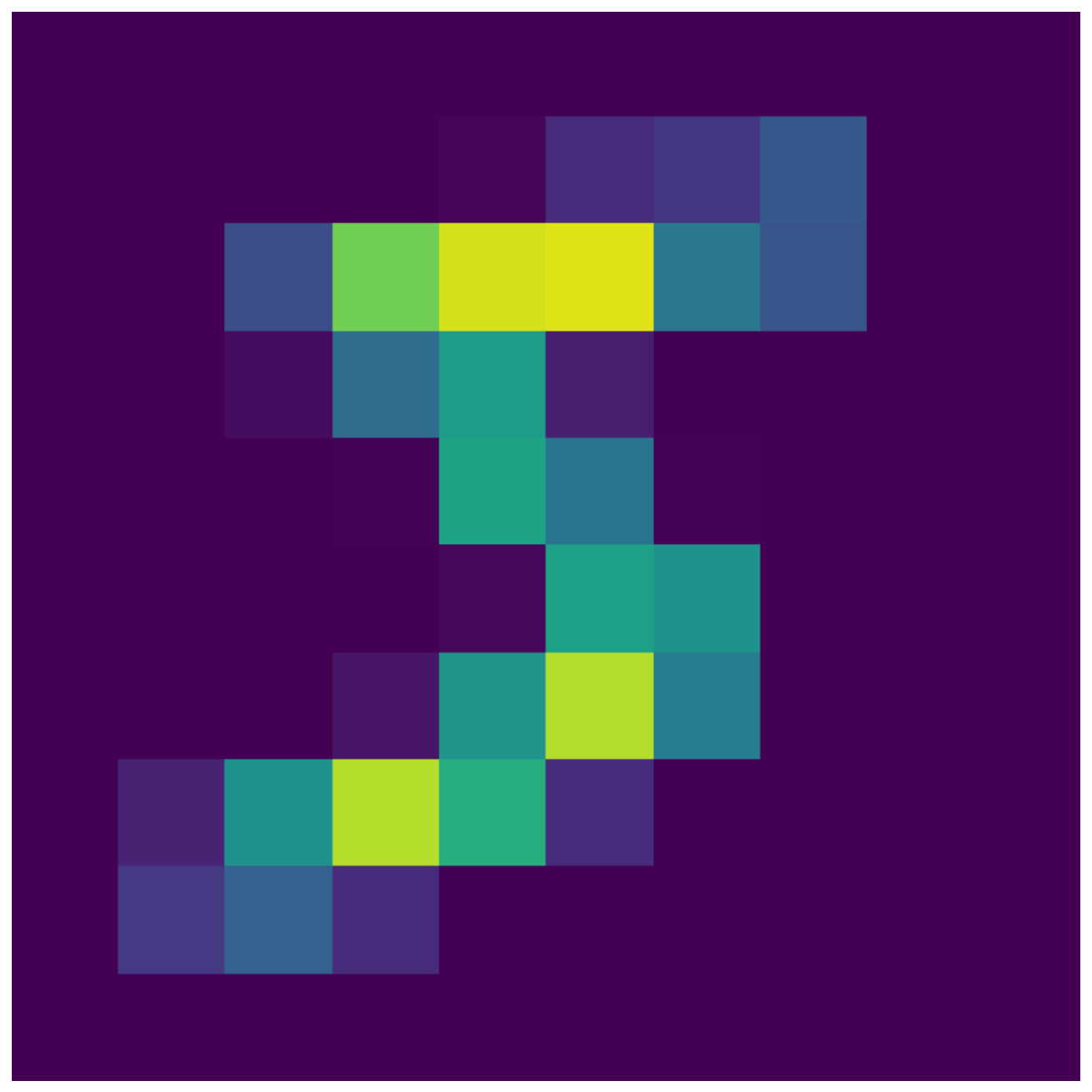}
    &\includegraphics[width=\imsize]{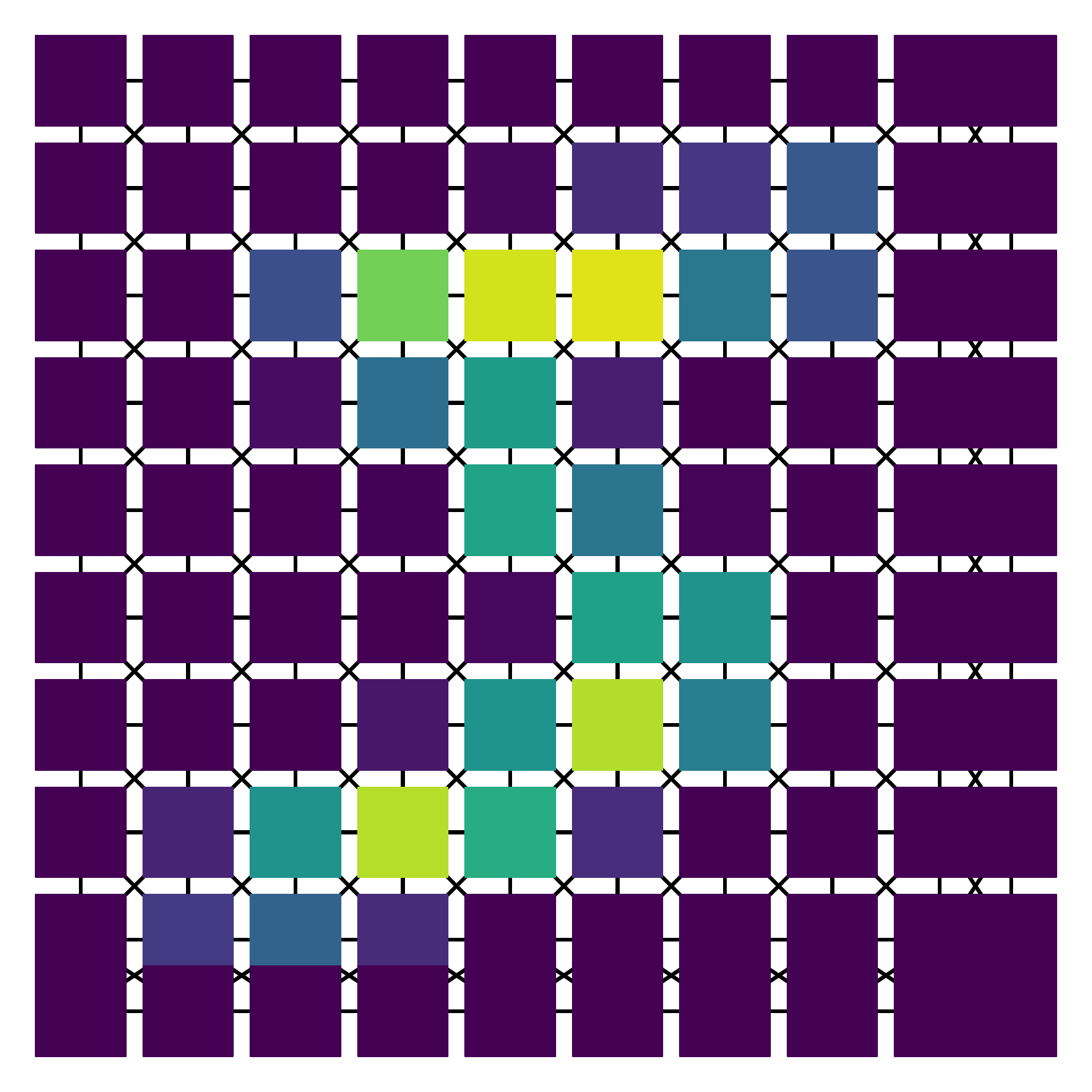}
    &\includegraphics[width=\imsize]{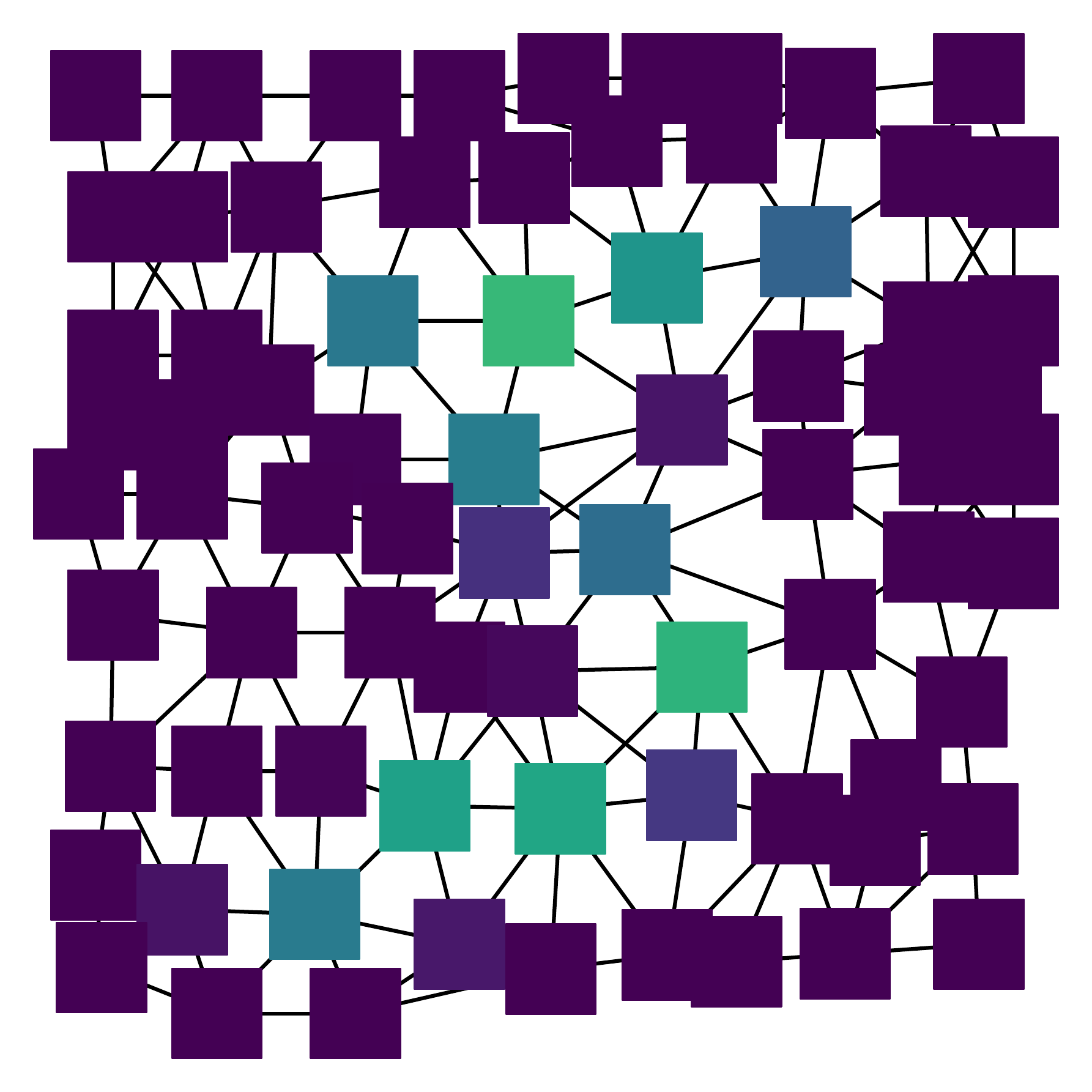}
    &\includegraphics[width=\imsize]{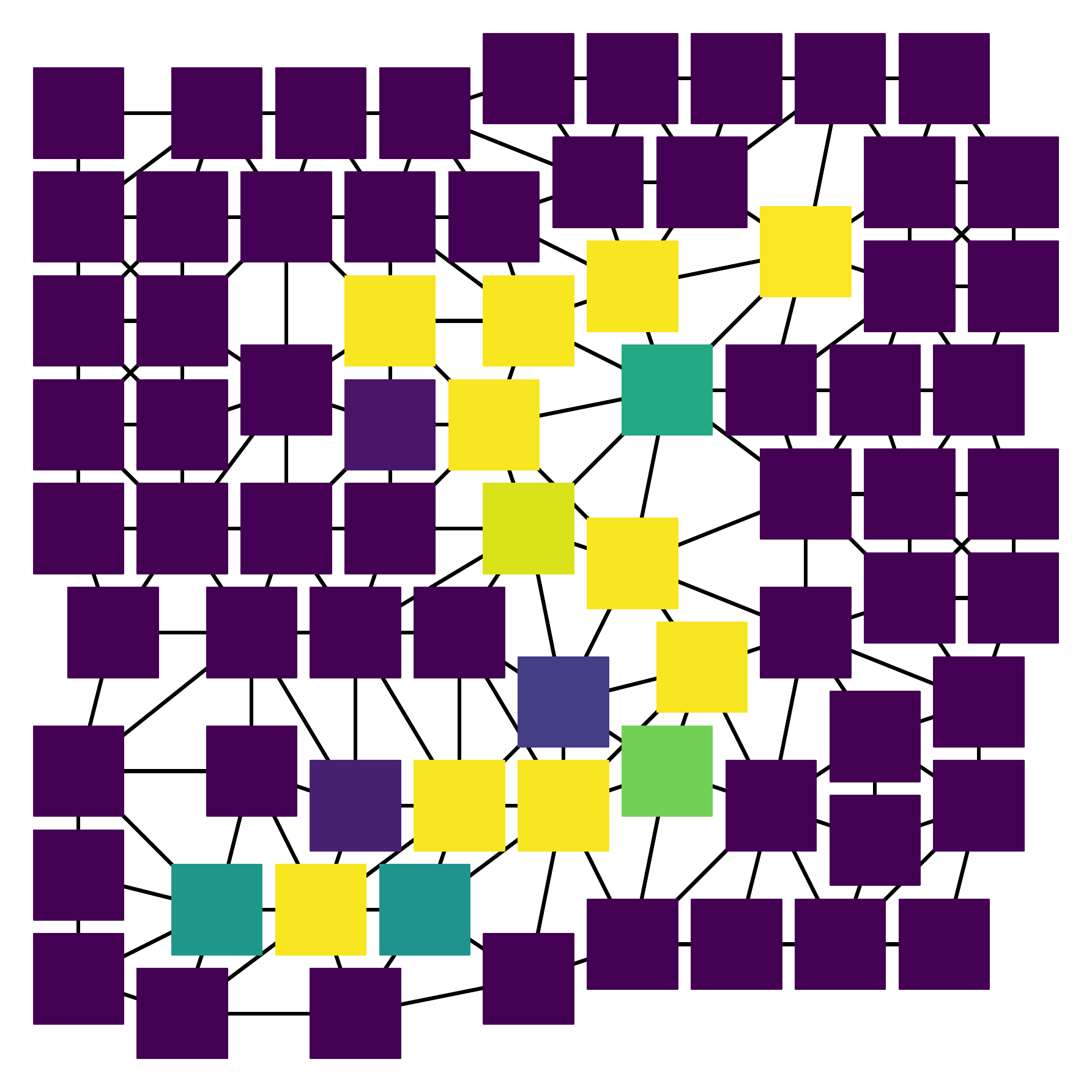}\\[-4pt]
    \rotatebox{90}{\small\quad\quad\ \ \ $k = 3$}
    &\includegraphics[width=\imsize]{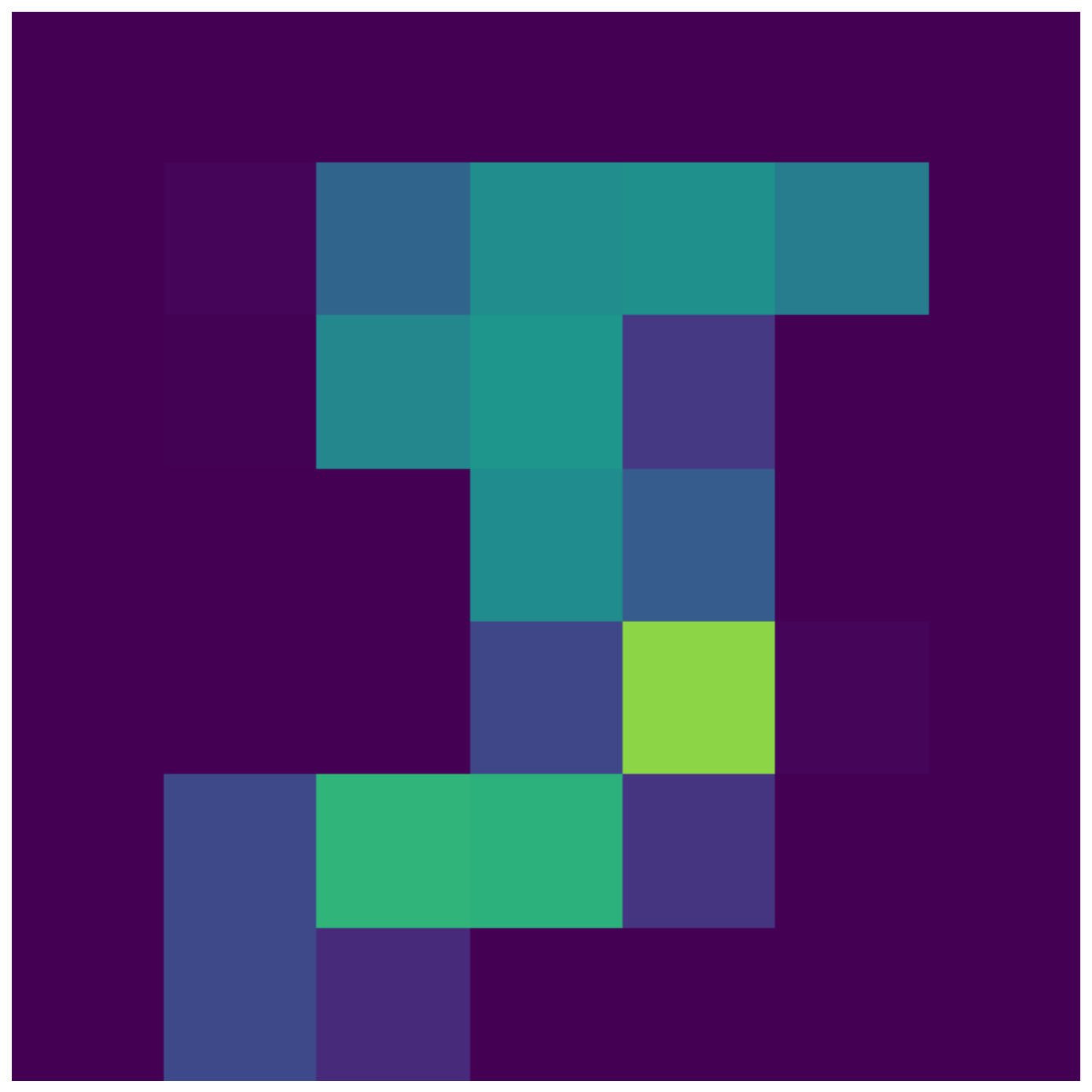}
    &\includegraphics[width=\imsize]{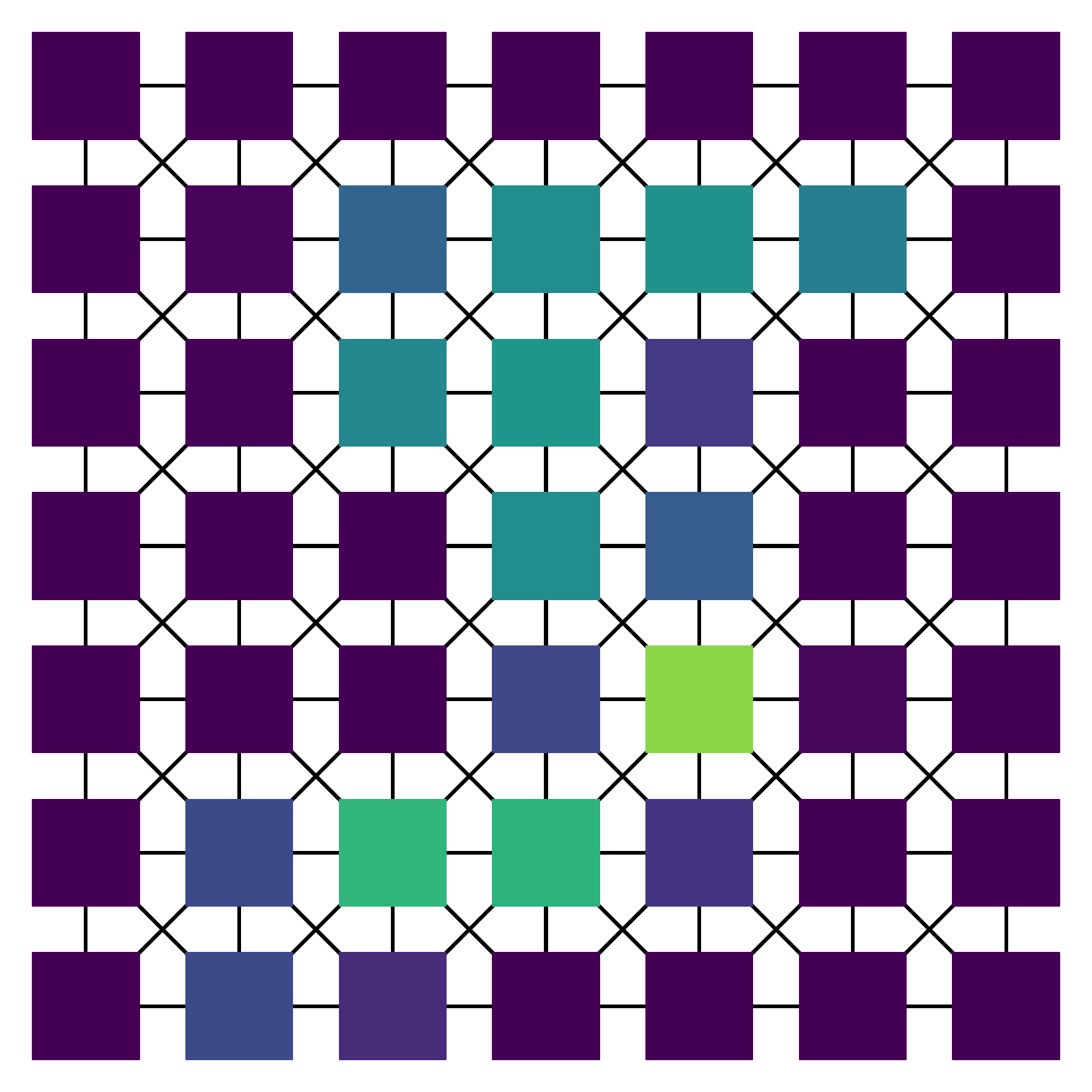}
    &\includegraphics[width=\imsize]{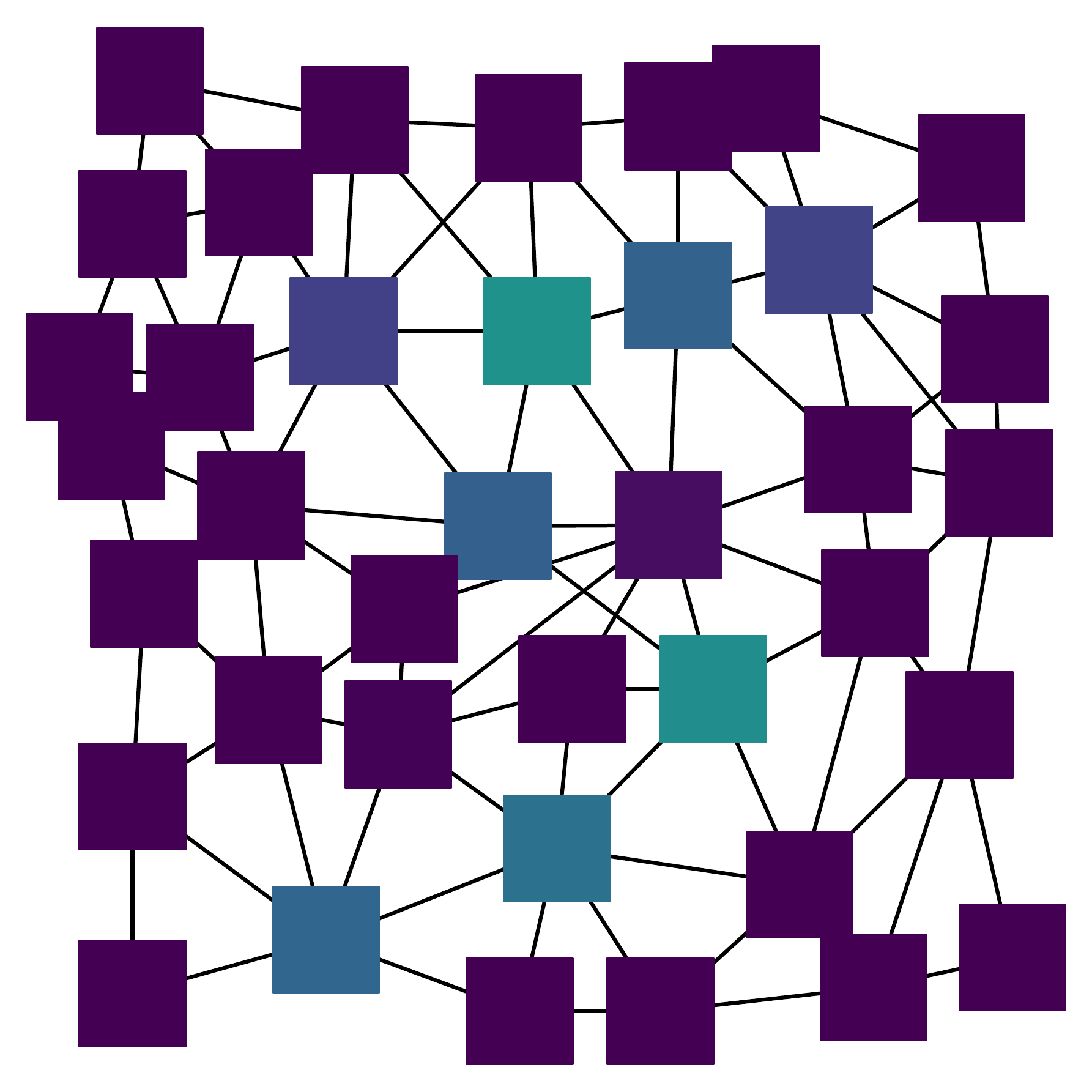}
    &\includegraphics[width=\imsize]{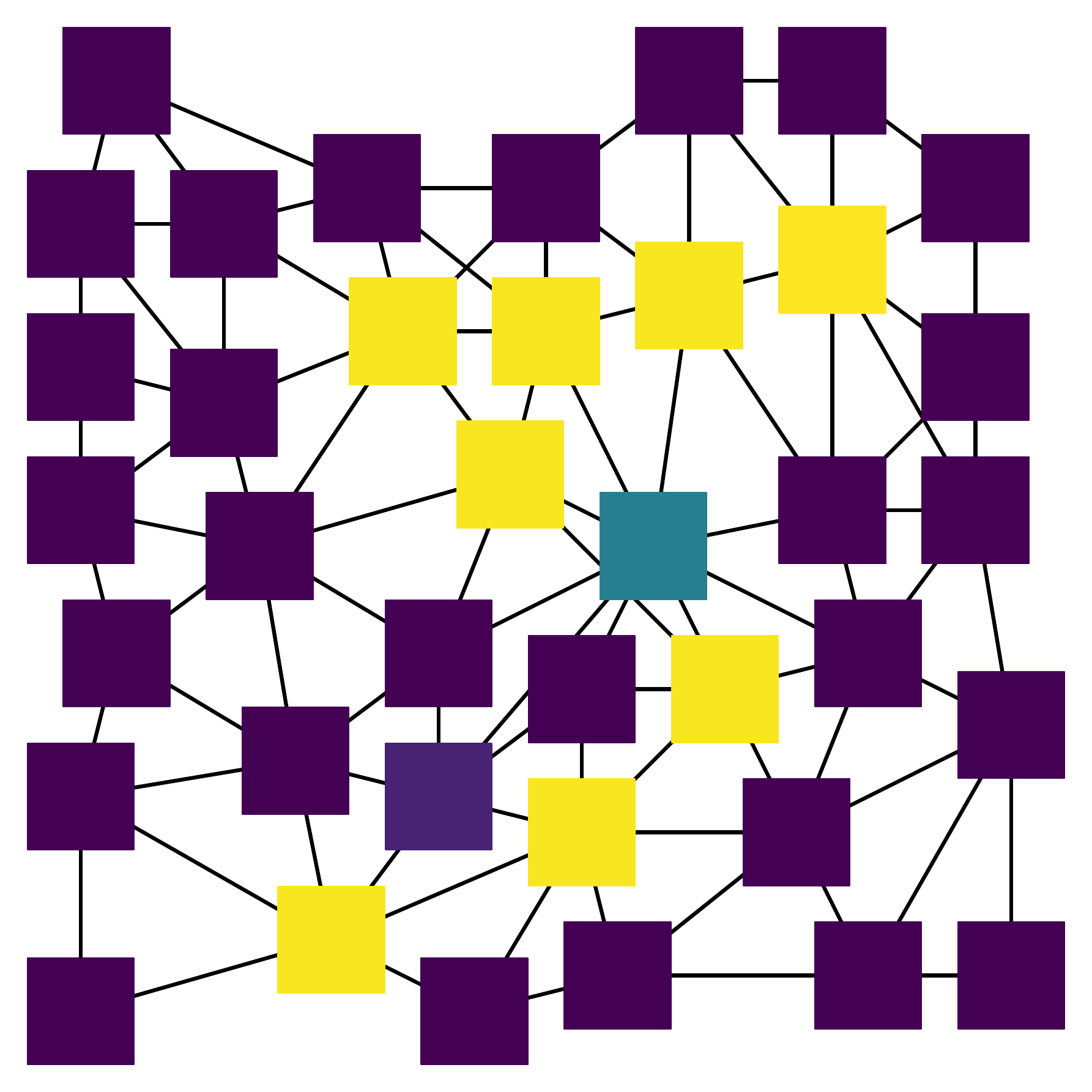}\\[-4pt]
    \rotatebox{90}{\small\quad\quad\ \ \ $k = 4$}
    &\includegraphics[width=\imsize]{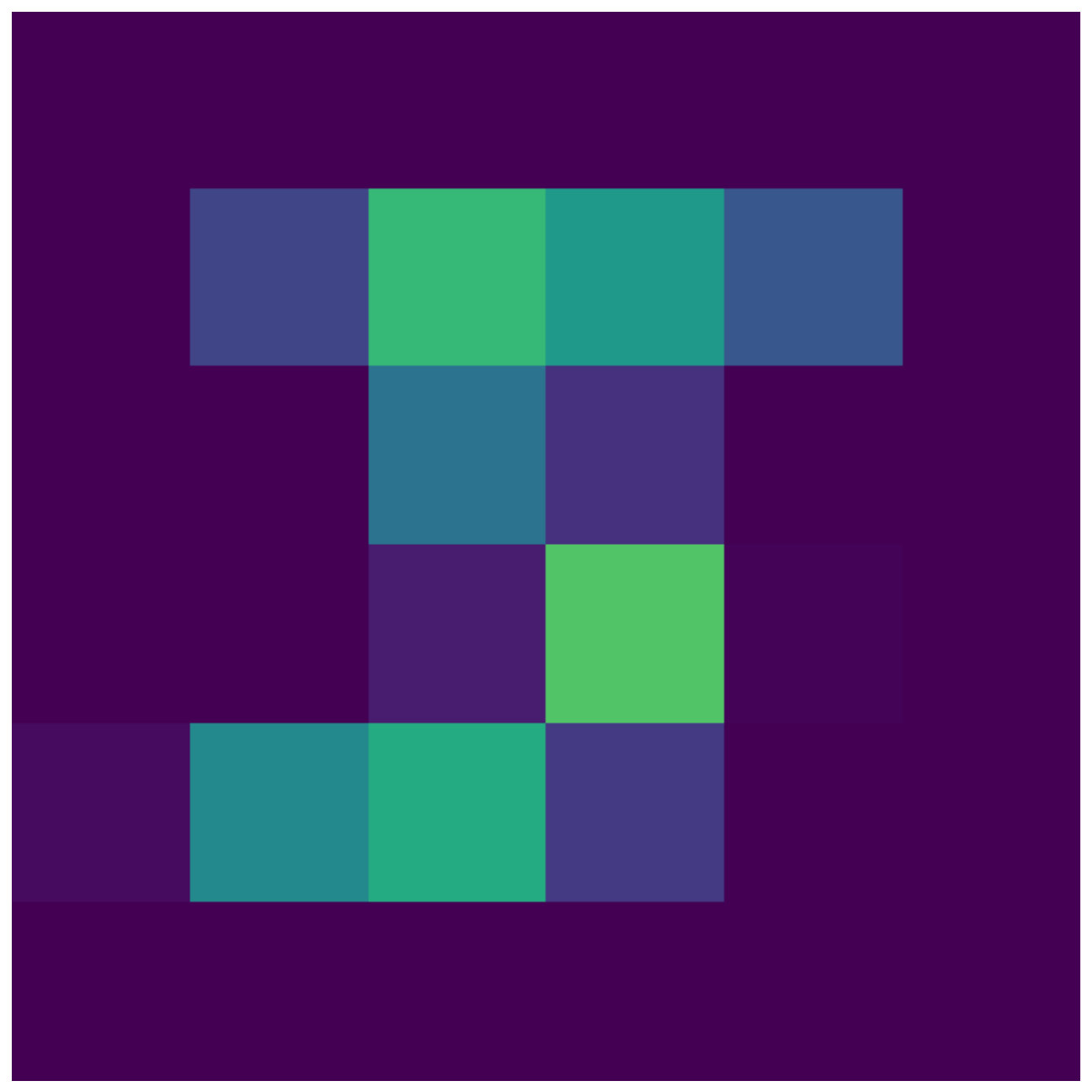}
    &\includegraphics[width=\imsize]{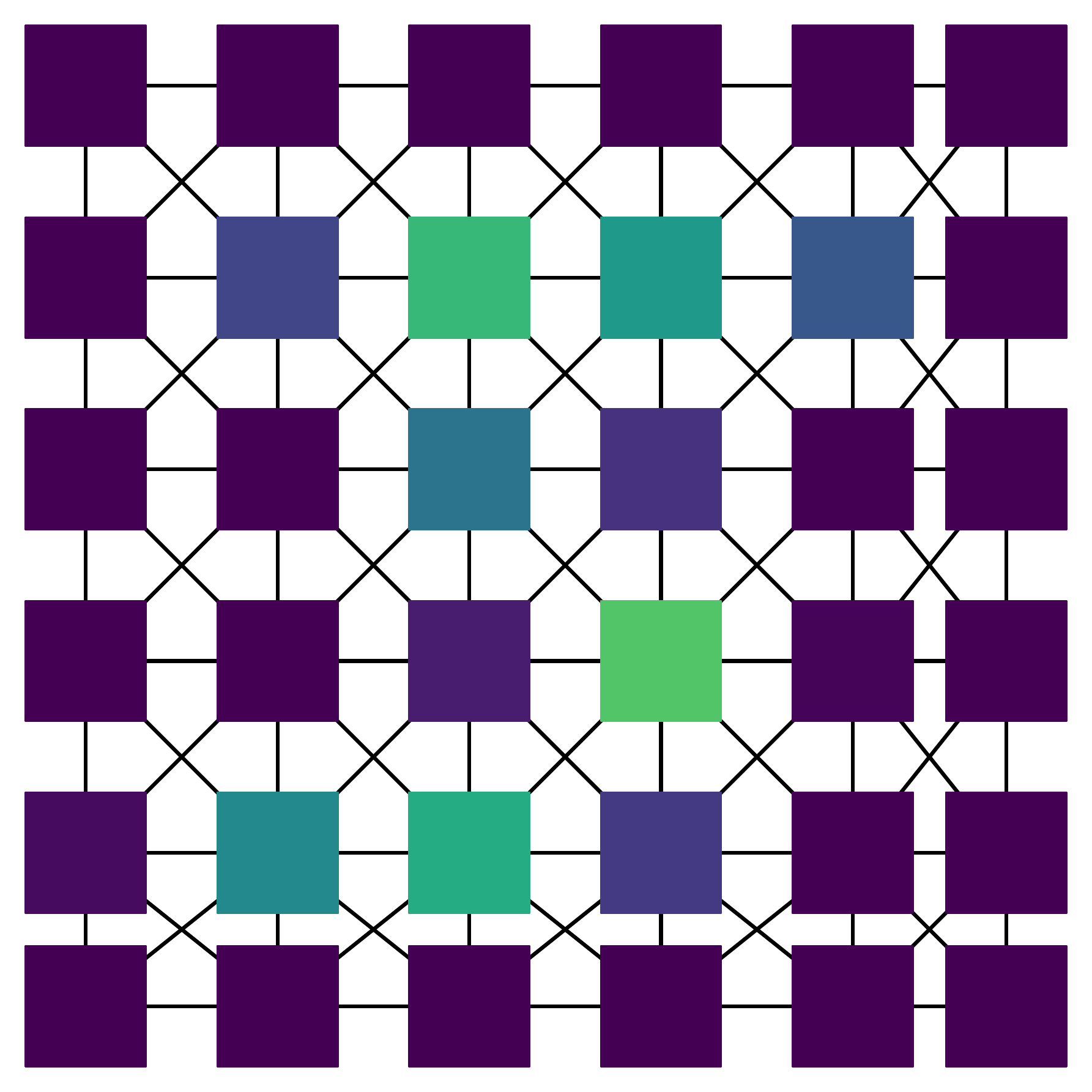}
    &\includegraphics[width=\imsize]{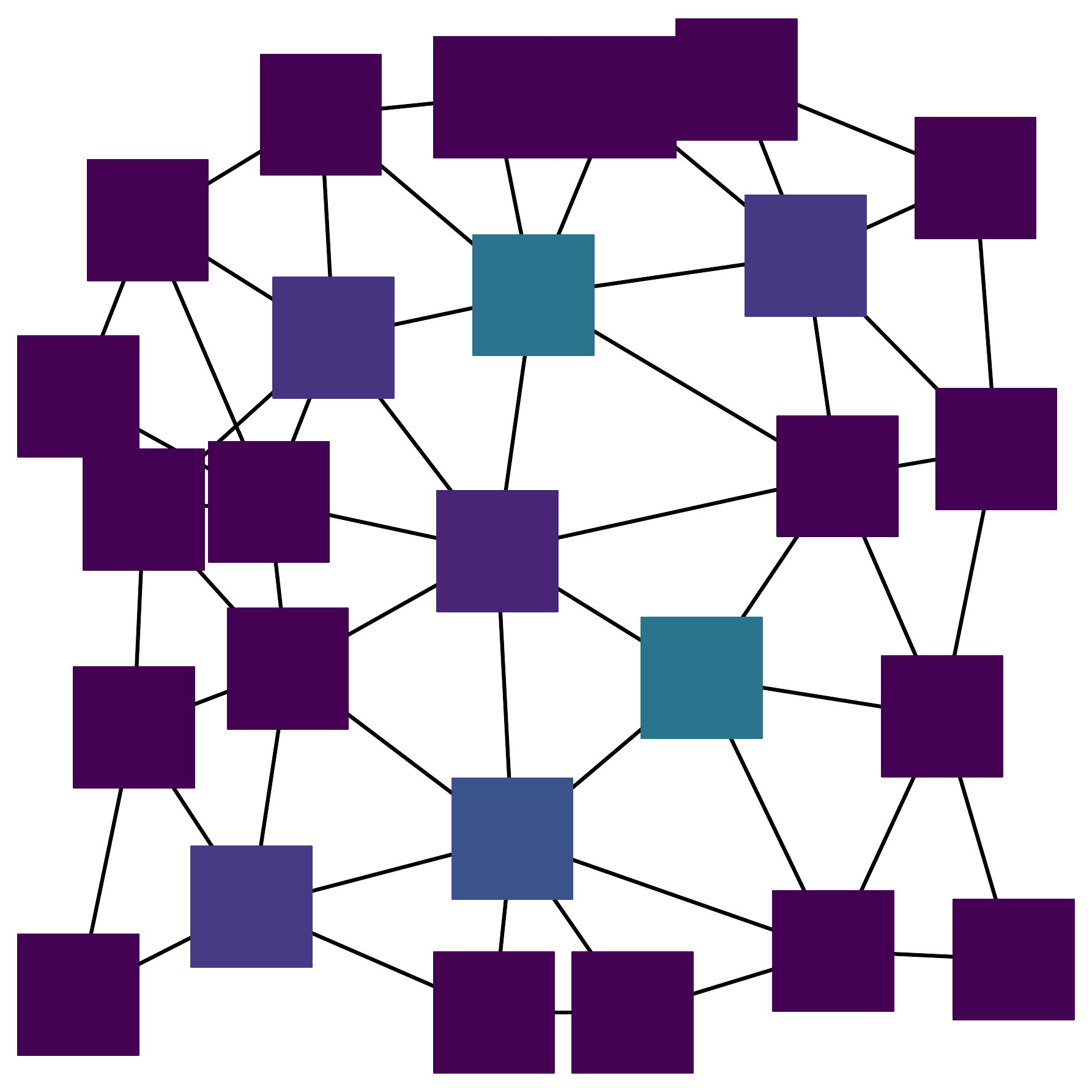}
    &\includegraphics[width=\imsize]{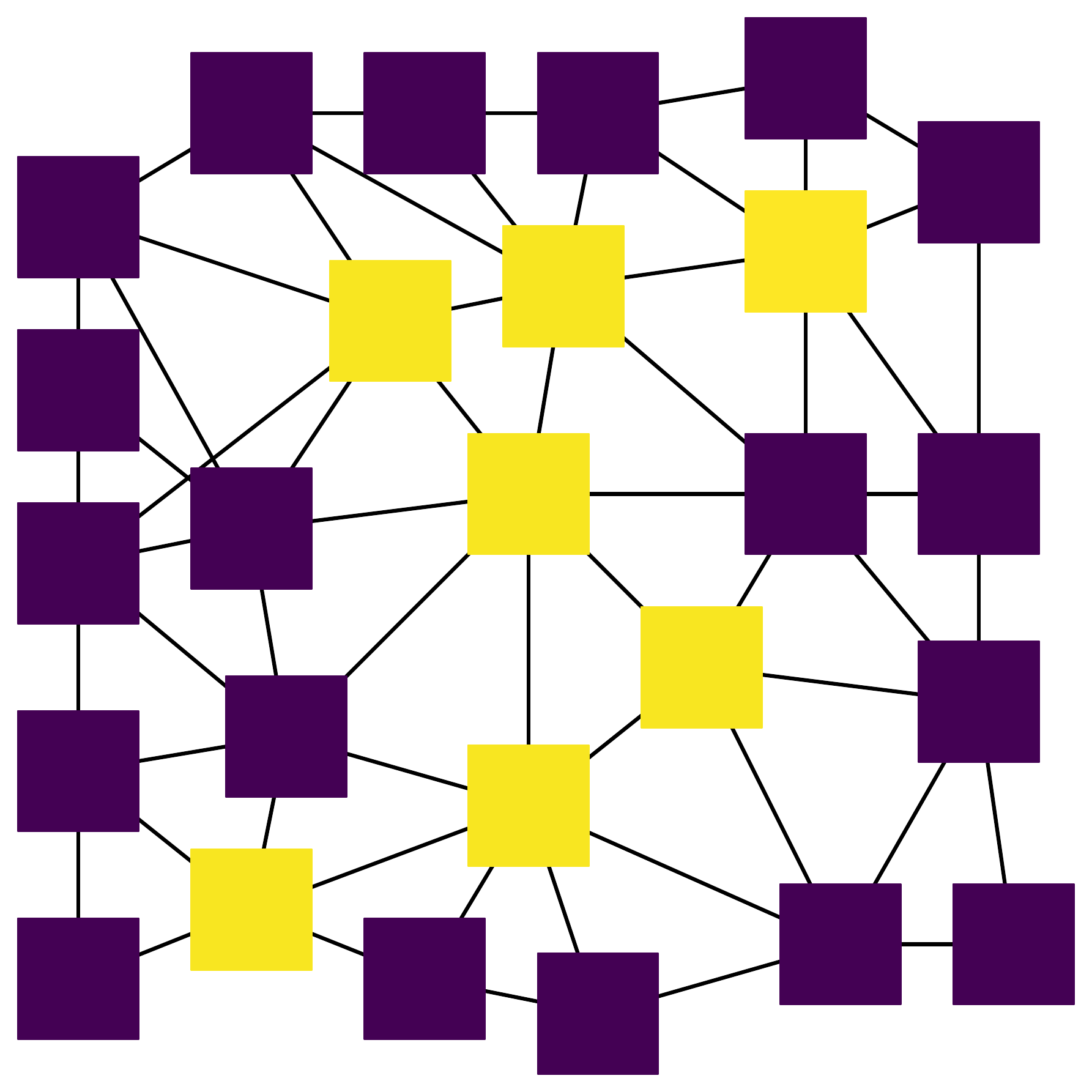}
    \end{tabular}}\vspace*{-0.75ex}
    \caption{\emph{(first column)} Average pooling and \emph{(second to fourth columns)} our method using different ranking and aggregation functions, for varying values of $k$.}
    \label{fig:mnist}
\end{figure}
\paragraph{Node ordering.}

A key property of our \kmis algorithm (similarly to the one of~\citet{blelloch_greedy_2012}) is that it is \emph{deterministic}: given a graph $\G$ and a \emph{ranking} of its nodes $\rank: V \to \{1,\dots,n\}$, that defines the position of the nodes in a given ordering, \cref{alg:k-mis} will always produce the same \kmis, for any $k \ge 0$.
This property has some interesting consequences:
%
\begin{itemize}
    \item The ranking $\rank$ can be used to lead \cref{alg:k-mis} to greedily include nodes having a higher rank under a given order of importance, such as a centrality measure, a task-dependent relevance, or a (possibly learned) \emph{scoring value}. (Note that the computation of the ranking can impact the complexity of the algorithm.) 
    \item If the ranking can be \emph{uniquely} determined by the nodes themselves (e.g., in function 
    of their attributes or 
    their neighbors), \cref{alg:k-mis,alg:cluster} become injective and hence, permutation invariant.\footnote{Notice that, in our setting, if $\rank: V \to \{1, \dots, n\}$ is injective, then it is also bijective and, hence, a permutation.} This can be obtained
    by ranking the nodes with respect to a score computed by means of a (sufficiently expressive) GCN, as learning injective functions over the nodes in a graph is a problem strictly related to the one of graph isomorphism, a topic that is gaining a lot of traction in the graph learning community~\cite{morris_weisfeiler_2019,xu_how_2019,maron_provably_2019,loukas_how_2020,geerts_expressiveness_2021,papp_theoretical_2022}. 
    \item A properly chosen ranking can produce a marginally greater total score of the selected nodes with respect to the one that we would get by greedily selecting the top scoring ones. This aspect 
    will be discussed more in detail in \cref{sect:theo}.
\end{itemize}

We now provide two examples on how we can change the ranking of the nodes to prioritize salient aspects according to a specific preference. Examples are conducted on the graph defined by the first sample of the MNIST dataset~\cite{lecun_mnist_2010}, a $28\times 28$ monochromatic image (first row of \cref{fig:mnist}) where every pixel is connected to the ones in the same pixel row, column or diagonal. 

First, we simulate the typical downsampling on images (also known as \emph{average pooling}~\cite{fukushima_neocognitron_1980}), where squared partitions of $p\times p$ pixels are averaged together (first column of \cref{fig:mnist}). To do this, we set the ranking $\rank$ of \cref{alg:cluster} as the \emph{lexicographic} ordering: given $(i, j)$ the coordinate of a pixel, we rank the nodes in decreasing order of $28i + j$.
The resulting reduction is in the second column of \cref{fig:mnist}: averaging intensities of pixels in the same partition produces a coarsened graph which is identical to classical downsampling.  
Note that this result is partly due to the fact that \cref{alg:cluster} also makes use of $\rank$ to define the clustering, such that the nodes in a partition have always a lower rank with respect to the centroid in the same partition. 

Secondly, we rank nodes in decreasing order of \emph{intensity}, thus prioritizing the pixels (i.e.,  the nodes) belonging to the drawn digit. Here we show two different results: the first, where we average the lightness and coordinates of the nodes in the same clusters (third column of \cref{fig:mnist}), and a second one, where we just keep the ones belonging to the nodes in the \kmis (fourth column). We see that the reduced graphs indeed prioritized the digit against other pixels, producing a coarsened representation where the digit is also remarkably recognizable. 



\section{Theoretical Analysis and Results}\label{sect:theo}


\paragraph{Regular downsampling.}

Downsampling plays a key role in Convolutional Neural Networks~\cite[CNNs,][]{goodfellow_deep_2016}, where it is adopted, for instance, in \emph{strided} convolutions and \emph{pooling} layers. In strided convolutions, an input tensor (e.g., a time series, an image, or a voxel grid) is reduced by applying the convolved filter every $s$-th of its entries, along every dimension, while skipping the others. 
In pooling layers, instead, tensors are reduced by summarizing every $p$-sided sub-tensors, taken at regular intervals. (More specific reductions are also possible, where distinct intervals are used for every dimension.)

We can show that, on $n$-dimensional 
diagonal grid graphs (i.e., grids where nodes are also diagonally adjacent), 
\cref{alg:k-mis,alg:cluster} behave \emph{exactly} as the aforementioned downsampling strategies, if we rank their nodes by their position in lexicographic order. This is of particular interest as the adjacencies in these graphs can represent the receptive fields of a single convolutional layer when applied to a some regular data of the same shape, like  
images (2-dimensional) or voxel grids (3-dimensional).
Specifically, if $\G = (V, E)$ is a diagonal grid constructed using the entries of a given tensor as nodes, and $\rank$ is the ranking of these entries in lexicographic order of their position indices, we have that
\begin{enumerate}
    \item $k\text{-MIS}(\G, V, \rank)$ selects the same entries of a strided convolution with $s=k+1$,
    \item $\textsc{Cluster}(\G, k, \rank)$ partitions the tensor as a pooling layer with $p = k+1$, and
    \item the reduced graph obtained by contracting the resulting partition is again a diagonal grid of the same dimensions of their output tensor.
\end{enumerate}

A formal restatement and proof of these properties are provided in the Supplementary Material, while in \cref{fig:mnist} we show an example of the equivalence between pooling (first column) and our reduction method (second column).

\paragraph{Connectivity of the reduced graph.}

For the sake of conciseness, hereafter we denote with $(\H, \rho) = \reduce(\G, k)$ the function reducing a graph  $\G$ by contracting the clusters obtained with \cref{alg:cluster}, as described in \cref{sect:kmis}. The term $\H = (S, E')$ denotes the reduced graph, where $S$ is the \kmis of $\G$, while $\rho: V \to S$ is the function mapping every node to the (exactly one) centroid in its cluster. The following results are invariant with respect to the ranking parameter and the aggregation function used to reduce the edges or the nodes.



\begin{figure*}[tb]
    \centering
    \def\svgwidth{\linewidth}
    \hspace{-2ex} 
    \resizebox{\linewidth}{!}{%
    \input{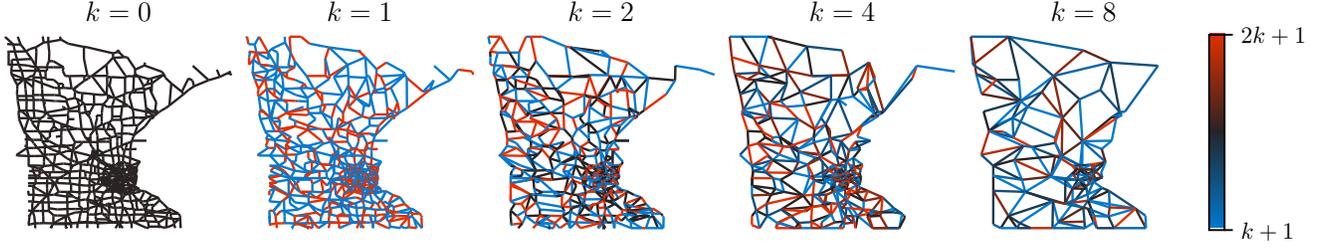}}%
    \vspace{-2ex} 
    \caption{Minnesota road network~\cite{davis_university_2011} reduced with different values of $k$. For $k = 0$, the two bounds coincide, as the graph is not reduced at all. For $k = 1$, the real distance covered by an edge is polarized (is either 2 or 3). For greater values of $k$, the edges' real distance span over all the range $[k +1, 2k + 1] \cap \nat$.}
    \label{fig:minnesota}
\end{figure*}

We follow a simple observation: for every edge in $uv \in E'$ with $u \neq v$, the nodes $u$ and $v$ are within $2k + 1$ hops in $\G$, since two nodes in $S$ are connected in $\H$ only if an edge in $\G$ crosses their two clusters. This property, combined with the lower bound implicitly defined by the \kmis, yields the following bounds.
\begin{remark}\label{rmk:centroid_hops}
For any $uv \in E(\H)$ such that $u \neq v$, we have that
$ k + 1 \le \length_\G(u, v) \le 2k + 1$.
\end{remark}
An example of this property is shown in \cref{fig:minnesota}, where bounds in \cref{rmk:centroid_hops} apply for the Minnesota road network~\cite{davis_university_2011} reduced with different values of $k$.
From the above observation, we can obtain the two following properties.
\begin{proposition}\label{thm:length}
	Let $\G$ be a connected graph and $(\H, \repr)  = \reduce\big(\G, k\big)$, with $k \ge 0$. Then, $\forall u, v \in V(\G)$,
	\[
	\length_{\H}(\repr(u), \repr(v)) \le \length_\G(u, v) \le (2k + 1)\length_{\H}(\repr(u), \repr(v)) + 2k.
	\]
\end{proposition}
\begin{corollary}\label{th:cc}
	For any $k\ge 0$, $\G$  and $\H = \mathcal{R}\big(\G, k\big)$ have the same number of connected components.
\end{corollary}
%
The full proofs are provided in
the Supplementary Material.
Both \cref{thm:length,th:cc} are fundamental in our proposal of using \kmis reduction as a pooling method in Graph Neural Networks. In particular:
\begin{enumerate*}[label=\emph{(\roman*)}]
    \item differently from several other pooling techniques~\cite{cangea_towards_2018,gao_graph_2019,knyazev_understanding_2019,lee_self-attention_2019,zhang_structure-feature_2020,ranjan_asap_2020,ma_path_2020}, we can guarantee that the input graph is not divided in multiple components, and that, if applied repeatedly, our method will eventually produce a single representation node for the whole graph;
    \item when training with batches of graphs at a time, our method guarantees also that different graphs are not joined together.
\end{enumerate*}

\paragraph{Algorithm discussion and complexity.} 

In order to avoid computing the $k$-th graph power of a possibly large-scale graph, \cref{alg:k-mis} modifies the one by \citet[][%
Algorithm~2, also restated in the Supplementary Material]{blelloch_greedy_2012} to compute the \kmis without explicitly generating every $k$-hop neighborhood. Given a graph $\G = (V, E)$, a subset of its nodes $U \subseteq V$, and a (injective) node mapping $\rank: V \to \{1, \dots, n\}$ (that we can consider as a {ranking} of the nodes under a given permutation), \Cref{alg:k-mis} works as follows:
\begin{enumerate}
    \item\label{item:step-1} if $U\subseteq V$ is not empty, in 
    \cref{line:k-hop-1,line:k-hop-2,line:k-hop-2-1,line:k-hop-2-2,line:k-hop-3}
    we find the set of nodes $S_0$ with minimum rank among their $k$-hop neighbors (i.e., their neighbors in $\G^k$). This is done with $k$ steps of label propagation such that, at each step, every node takes the minimum label found within their 
    ($1$-hop) neighbors. We only propagate labels belonging to nodes still in $U$;
    \item\label{item:step-2} in \cref{line:k-restrict-2,line:k-restrict-2-1,line:k-restrict-3} we remove from $U$ all the nodes that are at most $k$-hops from a node in $S_0$ (i.e., all their neighbors in $\G^k$). This is also done with $k$ steps of label propagation starting from the nodes in $S_0$, where this time the propagated label is a flag signaling that the node shall be removed;
    \item\label{item:step-3} finally, the algorithm makes a recursive call in \cref{line:k-mis-rec} using only the remaining nodes. The resulting set is merged with $S_0$ and returned.
\end{enumerate}
It is easy to see that, if $k=1$, steps \labelcref{item:step-1,item:step-2,item:step-3} become exactly Blelloch's algorithm,
whereas by taking a general $k$ every step is extended to consider $k$-hop neighbors of $\G$, thus efficiently emulating Blelloch's algorithm on $\G^k$. 



As for complexity, \citet{blelloch_greedy_2012} propose several trade-offs between \emph{work} and \emph{depth} on a \emph{concurrent-read/concurrent-write} PRAM model (CRCW, with minimum priority concurrent write). Here, we consider one version (Algorithm~2 from \citet{blelloch_greedy_2012})
which allows an efficient parallel implementation with $\bigO{m}$ work and $\bigO{\log^3 n}$ depth with high probability~\cite[see][Lemma~4.2]{blelloch_greedy_2012}, and most closely resembles the structure of \cref{alg:k-mis}.
Our algorithm introduces a factor $k$ (compared to
the one of \citet{blelloch_greedy_2012})
on the operations performed on lines~\cref{line:k-hop-1,line:k-hop-2,line:k-hop-2-1,line:k-hop-2-2,line:k-hop-3} and~\cref{line:k-restrict-2,line:k-restrict-2-1,line:k-restrict-3} to compute the $k$-hop neighborhood. It follows that the work and depth of \cref{alg:k-mis} are bounded by $k$ times that of Blelloch's algorithm, i.e., $\bigO{k(n+m)}$ work and $\bigO{k \log^3 n}$ depth w.h.p., where an extra $\bigO{n}$ work is needed to generate the additional vector of labels, which is modified every $k$ iterations. 
%
Regarding \cref{alg:cluster}, after computing the \kmis, the algorithm performs $k$ steps of label propagation, which add $\bigO{k(n+m)}$ work and $\bigO{k\log n}$ depth to the total computation. Total space consumption is $\bigO{n+m}$, comprising input and $\bigO{1}$ label vectors of size $\bigO{n}$. 
\begin{proposition}
Given a graph $\G$, an integer $k \in \nat$, and a random ranking of the nodes $\rank$, both \cref{alg:k-mis,alg:cluster} can be implemented to run on a CRCW PRAM using $\bigO{k(n + m)}$ work, $\bigO{k\log^3 n}$ depth, and $\bigO{n+m}$ space. The depth bound holds w.h.p.
\end{proposition}

\paragraph{Bounds on the total weight.}

In any greedy MIS algorithm, whenever we add a node to the independent set we have to remove all of its neighbors from the graph. Having observed this, a typical heuristic to compute larger-weight independent sets is to select nodes with high weight and low degree~\cite{caro_new_1979,wei_lower_1981}. Following this intuition, \citet{sakai_note_2003} proposed the following rules: given $\vec{x} \in \real_+^n$ a vector of positive weights associated to each node, add to the independent set the node $v$ maximizing either
\begin{enumerate*}[label=\emph{(\roman*)}]
    \item\label{item:sakai-1} $\vec{x}_v/(\degree(v) + 1)$, or
    \item\label{item:sakai-2} $\vec{x}_v/(\sum_{u \in \neigh[v]} \vec{x}_u)$.
\end{enumerate*}
Both rules can be trivially extended to $k$-hop neighborhoods by computing $\G^k$, which would however require $\bigO{n^2}$ space, unless done sequentially. Parallel computation of the \emph{neighborhood function} $\degree_k(v) = \lvert\neigh_k(v)\rvert$ in limited space can be achieved only by resorting to approximations, e.g. using Monte Carlo methods~\cite{cohen_size-estimation_1997} or approximate sets representations~\cite{palmer_anf_2002,boldi_hyperanf_2011}, and still this would not extend to approximate rule~\ref{item:sakai-2}.

To overcome these limitations, we 
overestimate the sum of the weights in the $k$-hop neighborhood of each node, by computing instead $\vec{c}_k = (\adj + \I)^k\vec{x} \in \real_+^n$, where $\adj, \I \in \{0, 1\}^{n\times n}$ are, respectively, the adjacency and the identity matrices. The matrix $(\adj + \I)^k \in \nat_0^{n\times n}$ represents the number of $k$-walks (i.e., sequences of adjacent nodes of length $k$) from every pair of nodes in the graph. Clearly, $[(\adj + \I)^k]_{uv} \ge 1$ if $v \in \neigh_k[u]$, while the equality holds for every pair of nodes for $k=1$. When $\vec{x} = \vec{1}$, $\vec{c}_k$ is equal to the $k$-path centrality~\cite{sade_sociometrics_1989,borgatti_graph-theoretic_2006}. Notice that we do not need to compute $(\adj + \I)^k$ explicitly, as $\vec{c}_k$ can be obtained with a sequence of $k$ matrix-vector products, that can be computed in $\bigO{n+m}$ space, $\bigO{k(n+m)}$ work and $\bigO{k\log n}$ depth. 

In the following, we provide a generalization of the bounds of \citet{sakai_note_2003} when a \kmis is computed by \cref{alg:k-mis} with the ranking defined by rules \ref{item:sakai-1}-\ref{item:sakai-2} approximated by the $k$-walk matrix $\vec{c}_k$. We remark that, for $k=1$, the following theorems are providing the same bounds as the one given by \citet{sakai_note_2003}. The full proofs can be found in
the Supplementary Material.


\begin{theorem}\label{th:bound-k-degree}
Let $\G = (V, E)$ be a graph, with (unweighted) adjacency matrix $\adj \in \{0, 1 \}^{n\times n}$ and with $\vec{x} \in \real_+^n$ representing a vector of positive node weights. Let $k \in \nat$ be an integer, then define $w: V \to \real_+$ as
\begin{align}\label{eq:rank-k-degree}
    w(v) = \frac{\vec{x}_v}{[(\adj + \I)^k\vec{1}]_v},
\end{align}
and $\rank_{w}$ as the ranking of the nodes in decreasing order of $w$. Then, $k\textnormal{-MIS}(\G, V, \rank_{w})$ outputs a maximal $k$-independent set $S$ such that $\sum_{u \in S} \vec{x}_u \ge \sum_{v \in V} w(v)$.
\end{theorem}

\begin{theorem}\label{th:bound-k-weights}
Let $\G = (V, E)$ be a graph, with (unweighted) adjacency matrix $\adj \in \{0, 1 \}^{n\times n}$ and with $\vec{x} \in \real_+^n$ representing a vector of positive node weights. Let $k \in \nat$ be an integer, then define $w: V \to \real_+$ as
\begin{align}\label{eq:rank-k-weights}
    w(v) = \displaystyle\frac{\vec{x}_v}{[(\adj + \I)^k\vec{x}]_v},
\end{align}
and $\rank_{w}$ as the ranking of the nodes in decreasing order of $w$. Then, $k\textnormal{-MIS}(\G, V, \rank_{w})$ outputs a maximal $k$-independent set $S$ such that $\sum_{u \in S} \vec{x}_u \ge \sum_{v \in V} w(v)\cdot\vec{x}_v$.
\end{theorem}%

\begin{theorem}\label{th:ratio}
Let $\G = (V, E)$ be a non-empty graph with positive node weights $\vec{x} \in \real_+^n$, and let $\rank_w$ be a ranking defined as in \cref{th:bound-k-degree} or \ref{th:bound-k-weights} for any given $k\in \nat$. Then
%
%
$
\sum_{v\in S} \vec{x}_v \ge \alpha(\G^k)/\Delta_k,
$
where $S = k\textnormal{-MIS}(\G, V, \rank_w)$ and $\Delta_k = \max_{v \in V}\ [(\adj + \I)^k\vec{1}]_v$.
\end{theorem}

Recalling that $\alpha(\G^k)$ is the optimal solution, \cref{th:ratio} shows that our heuristics guarantee a $\Delta_k^{-1}$ approximation. 
This bound degrades very quickly as the value of $k$ increases, since the number of $k$-walks may exceed the total number of nodes in the graph. In the Supplementary Material we show that, in practice, the total weight produced by \cref{alg:k-mis} is on par with respect to the one obtained using the exact neighborhood function for low values of $k$. This aspect is of practical value as in general the $k$ values used for graph pooling are on the low-end.

\section{Related Works} \label{sect:relate}

\paragraph{Maximal $\bm{k}$-Independent Sets.} 
Computing 
a \kmis can be trivially done in (superlinear) polynomial time and space using matrix powers~\cite{agnarsson_powers_2003} and any greedy MIS algorithm~\cite[e.g.,][]{luby_simple_1985,blelloch_greedy_2012}. 
\citet{koerts_k-independent_2021} proposed a formulation of the problem both as an integer linear program and as a semi-definite program, but still relying on the $k$-th power of the input graph. 
Several papers propose efficient algorithms to solve the {maximum} (weighted or unweighted) $k$-IS problem on specific classes of graphs~\cite{agnarsson_powers_2000,agnarsson_powers_2003,eto_distance-d_2014,bhattacharya_generalized_1999,duckworth_large_2003,pal_sequential_1996,hsiao_efficient_1992,hota_efficient_2001,saha_maximum_2003}, which fall beyond the scope of this article. 
To the best of our knowledge, the only other parallel algorithm for computing a maximal $k$-independent set was proposed by \citet{bell_exposing_2012} as a generalization of the one of \citet{luby_simple_1985} for $k \ge 1$. This algorithm is essentially the same as \cref{alg:k-mis,alg:cluster}, but without the ranking argument, making the algorithm non-deterministic, as the nodes are always extracted in a random order. 

\paragraph{Graph Coarsening and Reduction.} 
%
%
MISs (i.e., with $k=1$) were adopted as a first sampling step in \citet{barnard_fast_1994}, although their final reduction step may not preserve the connectivity of the graph.
Using MIS was also suggested by \citet{shuman_multiscale_2016} as an alternative sampling step for their graph reduction method. The spectral reduction proposed by \citet[neighborhood variant]{loukas_graph_2019} does not use sampling as a first reduction step, but sequentially contracts node neighborhoods until a halting condition is reached, performing similar steps to the classical greedy algorithm for maximum-weight independent sets.
\paragraph{Graph Pooling.} In a contemporary and independent 
work, \citet{stanovic_maximal_2022} introduced a pooling mechanism based on maximal independent (vertex) sets, named MIVSPool. Their method is analogous to ours, but restricted to the case of \kmis[1], that they compute using the parallel algorithm of \citet{meer_stochastic_1989}. Another related model is \textsc{EdgePool}~\cite{diehl_towards_2019}, which computes instead a maximal matching, i.e., a maximal independent set of edges, selecting the edges depending on a learned scoring function.
\citet{nouranizadeh_maximum_2021} also proposed a pooling method that constructs an independent set maximizing the mutual information between the original and the reduced graph. To do so, the authors leverage on a sequential algorithm with cubic time complexity and also no guarantees that the resulting set is maximal. 
Apart from a few other cases~\cite{dhillon_weighted_2007,luzhnica_clique_2019,ma_graph_2019,wang_haar_2020,bacciu_non-negative_2019,bacciu_k-plex_2021,bianchi_hierarchical_2020}, pooling in Graph Neural Networks (GNNs) usually entails an adaptive approach, typically realized by means of another neural network. These pooling methods can be divided in two types: \emph{dense} and \emph{sparse}. Dense methods, such as \textsc{DiffPool}~\cite{ying_hierarchical_2018}, \textsc{MinCutPool}~\cite{bianchi_spectral_2020,bianchi_hierarchical_2020}, \textsc{MemPool}~\cite{khasahmadi_memory-based_2019}, \textsc{StructPool}~\cite{yuan_structpool_2019}, and \textsc{DMoN}~\cite{tsitsulin_graph_2022}, compute for each node a soft-assignment to a fixed number of clusters defined by a reduction factor $r \in (0, 1)$, thus generating a matrix requiring $\bigO{rn^2}$ space. Sparse methods, such as \textsc{gPool/TopKPool}~\cite{gao_graph_2019,cangea_towards_2018}, \textsc{SAGPool}~\cite{lee_self-attention_2019,knyazev_understanding_2019}, \textsc{GSAPool}~\cite{zhang_structure-feature_2020}, \textsc{ASAPool}~\cite{ranjan_asap_2020},  \textsc{PANPool}~\cite{ma_path_2020}, \textsc{iPool}~\cite{gao_ipoolinformation-based_2021}, and \textsc{TAGPool}~\cite{gao_topology-aware_2021}, instead, compute a score for each node (requiring $\bigO{n}$ space), and reduce the graph by keeping only the top $\lceil rn \rceil$ scoring ones and dropping the rest. Although scalable, these methods provide no theoretical guarantees regarding the preservation of connectivity of the reduced graph, as the $n - \lceil rn \rceil$ dropped nodes may disconnect the graph. 
%

\section{Experimental Analysis} \label{sect:exp}

\begin{table*}[tb]
    \centering
    \small
\begin{tabular}{lr@{$\,\pm\,$}lr@{$\,\pm\,$}lr@{$\,\pm\,$}lr@{$\,\pm\,$}lr@{$\,\pm\,$}l}
\toprule
Model & \multicolumn{2}{c}{DD} & \multicolumn{2}{c}{REDDIT-B} & \multicolumn{2}{c}{REDDIT-5K} & \multicolumn{2}{c}{REDDIT-12K} & \multicolumn{2}{c}{GITHUB} \\
\midrule
\model{Baseline}           &    75.51 & 1.07 &         78.40 & 8.68 &       \ \,    48.32 &  2.38 &          \   45.04 & 6.63 &  {\bfseries 69.89} & {\bfseries 0.28} \\
\model{BDO}                &    \textit{\bfseries 76.69} & \textit{\bfseries 1.79} &         85.63 & 1.43 &           45.95 &  5.49 &            41.89 & 7.14 &  65.64 & 0.90 \\ 
\model{Graclus}            &    75.17 & 2.11 &         84.05 & 5.81 &           43.22 & 12.24 &            43.08 & 9.32 &  67.64 & 0.57 \\ 
\model{EdgePool}           &    74.70 & 1.57 &         85.98 & 1.57 &           52.44 &  1.11 &            {\bfseries 47.58} & {\bfseries 0.78} &  \textit{\bfseries 68.72} & \textit{\bfseries 0.52} \\ 
\model{TopKPool}           &    74.92 & 2.03 &         81.10 & 3.82 &           45.28 &  3.88 &            38.55 & 2.35 &  65.93 & 0.45\\ 
\model{SAGPool}            &    73.26 & 2.26 &         84.90 & 3.94 &           46.29 &  5.61 &            42.30 & 3.70 &  64.29 & 5.70\\ 
\model{ASAPool}            &    73.73 & 2.18 &         78.37 & 5.22 &           39.53 &  7.76 &            39.14 & 3.58 &  66.98 & 0.96\\ 
\model{PANPool}            &    73.26 & 1.94 &         77.44 & 4.95 &           46.04 &  3.78 &            40.97 & 3.02 &  62.48 & 2.84\\\midrule 
\kmis \emph{(strided)}      &    76.44 & 1.50 &         86.32 & 1.90 &           {\bfseries 54.30} &  {\bfseries 0.53} &            46.06 & 0.58 &  67.87 & 0.48 \\ 
\kmis \emph{(max-pool)}     &    {\bfseries 76.91} & {\bfseries 1.06} &         {\bfseries 87.57} & {\bfseries 1.96} &           53.44 &  1.52 &            \textit{\bfseries 47.51} & \textit{\bfseries 0.99} &  68.24 & 0.94 \\ 
\kmis \emph{(mean-pool)}    &    73.56 & 1.19 &         \textit{\bfseries 86.98} & \textit{\bfseries 1.13} &           \textit{\bfseries 54.00} &  \textit{\bfseries 0.94} &            46.73 & 0.85 &  68.60 & 0.67 \\ 
\bottomrule
\end{tabular}%

\caption{Classification accuracy on selected benchmark datasets \emph{(mean{\footnotesize $\bm{\,\pm\,}$}std)}}
\label{tab:accuracy}
\end{table*}
\cref{tab:accuracy} summarizes the average classification accuracy obtained on selected classification benchmarks using the same underlying Graph Neural Networks (GNNs) and different kinds of pooling mechanisms. For classification tasks, we chose those datasets having the highest number of nodes from in the \emph{TUDataset}~\cite{morris_tudataset_2020} collection
(i.e., DD~\cite{dobson_distinguishing_2003}, GITHUB-STARGAZERS~\cite{rozemberczki_karate_2020}, REDDIT-BINARY and -MULTI-5K/12K~\cite{yanardag_deep_2015}), 
where pooling layers may prove more useful. All datasets were divided in \emph{training} (70\%), \emph{validation} (10\%), and \emph{test} (20\%) sets using a randomized stratified split with fixed seed. All models have the same general architecture: 3 GNNs (optionally) interleaved by 2 layers of pooling, a global pooling method (\emph{sum} and \emph{max}), and a final MLP with dropout~\cite{srivastava_dropout_2014} as classifier. All models were trained using Adam optimizer~\cite{kingma_adam_2017}. We performed a model selection using the training and validation split, and then we computed the average test set classification accuracy obtained by the best configuration, on 10 inference runs using different seed values. 
The hyper-parameter concerning the reduction factor ($k$ in our case, or $r$ for other methods) has been chosen among the other parameters during the model selection phase. 
All models have been implemented and trained using PyTorch~\cite{paszke_pytorch_2019} and PyTorch Geometric~\cite{fey_fast_2019}.
A detailed description of the datasets, models, and experimental settings are provided in
the Supplementary Material, together with additional experiments regarding the controlled scaling and efficiency of our method, showing that it can reduce 
graphs with over 100 million edges in less than 10 seconds.
%

We compared our reduction method against different kinds of pooling layers readily available on the PyG library (we avoided \textsc{DiffPool}-like dense methods as they do not scale well on the selected datasets), and also against our own method using random rankings of the nodes, 
as done in the aggregation scheme of \citet[BDO,][]{bell_exposing_2012}. This method, along with the baseline (with no pooling) and \textsc{Graclus}, are the only compared architectures that require no additional parameters.
For our method, the node scoring function is computed by means of a sigmoid-activated linear layer having as input the features of the nodes. 
As in the other parametric methods, the feature vectors of the nodes are multiplied by the final score beforehand, 
to make the scoring function end-to-end trainable. 
The computed scores constitute the node weights and, as described in \cref{sect:theo}, the resulting ranking is obtained according to \cref{eq:rank-k-weights}. 
The reduction is performed in two settings: \emph{strided}, in which we perform no aggregation,
and \emph{pool}, in which we aggregate the feature vectors in each partition using, respectively, \emph{max} and \emph{mean} 
aggregations.


Looking at \cref{tab:accuracy} (where the top two accuracy scores for each dataset are in boldface) it is immediately evident how 
the proposed \kmis-based approaches obtain consistently high accuracy, suggesting that the
\emph{evenly-spaced} 
centroid selection is indeed able to capture essential properties of each graph. 
On the other hand, the random permutation variant of \citet{bell_exposing_2012},
seem to overall perform worse than the other \kmis-based strategies, while still obtaining a considerable result on DD and REDDIT-B. This suggests that exploiting the ranking function of \cref{alg:k-mis} to select relevant nodes is indeed able to improve the representativeness of the downsampled graph.
It is also particularly noteworthy how one of the best performing model, \textsc{EdgePool}, is also the only other parametric pooling method to preserve the connectivity of the graph, as its reduction step consists of contracting a maximal matching.
This highlights the importance of preserving the connectivity of the network when performing pooling in GNNs, while also suggesting that evenly-spaced reductions can benefit graph representation learning tasks.
%
Finally, we observe a remarkable performance of the baseline algorithm (no pooling) on the GITHUB
dataset: we may speculate that the graphs are simple enough to not require pooling, yet at the same time \kmis approaches obtains competitive accuracy, suggesting it is a reliable and versatile choice.


\section{Conclusions}

We introduced a new general graph coarsening approach that
aims
to preserve fundamental topological properties of the original graph, acting like a structured counterpart of downsampling methods for regular data.
The coarsening reduction can be regulated by the parameter $k$, going from the original graph, when $k=0$, to up to a single node as $k$ approaches the graph's diameter, shrinking graphs uniformly as pairwise distances maintain a stretch controlled by $k$. 
Furthermore, we showed how this parameter generalizes to the pooling and stride intervals when applied to diagonal grid graphs.

The algorithm is designed to provide such guarantees 
while at the same time allowing a scalable parallel implementation, which processes graphs with up to 100 million edges in just a few seconds on a single GPU. 

The empirical analysis provided evidence of effectiveness of our \kmis pooling in several graph classification benchmarks, showing superior performance with respect to related parametric and non-parametric methods from the literature. 


This approach fills a methodological gap between reduction techniques for structured data and their rigorous counterparts on regular data. Given its generality and scalability, it has potential of positively impacting a plethora of com\-pu\-ta\-tion\-al\-ly-intense applications for large scale networks, such as graph visualization, 3D mesh simplification, and classification. 


\begin{small}
\bibliography{bibliography.bib}
\end{small}

%
%
%
%
%
%
%
%
%
%
%
%
%
%
%
%
%
%
%
%
%
%
%
%
%
%
%
%
%
%
%
%


\iftoggle{arxiv}{
\vfill\break
\appendix

\begin{center}
    \LARGE\bf 
Supplementary Material
\end{center}

\section{Blelloch's Algorithm}\label{sec:mis-alg}

In \cref{alg:seq-mis,alg:mis} we restate, respectively, the sequential and its equivalent parallel greedy MIS algorithm, as proposed by \citet[Algorithm~1 and 2]{blelloch_greedy_2012}.
\begin{table}[h]
\setlength{\intextsep}{0pt}  
\begin{algorithm}[H]
\begin{algorithmic}[1]
	\Function{MIS}{$\G = (V, E)$, $\rank$}
	\State\algorithmicif\ $\lvert V \rvert = 0$ \algorithmicthen\ \Return $\emptyset$
	\State $v \gets \operatorname{argmin}_{u \in V} \rank(u)$
	\State $R \gets V \setminus \neigh[v]$
	\State\Return $\{v\} \cup {}$\Call{MIS}{$\G[R]$, $\rank$}
	\EndFunction 
\end{algorithmic}
\caption{Sequential Greedy \textsc{MIS} algorithm, from~\citet{blelloch_greedy_2012}. Given a graph $\G$ and a node ranking $\rank$, returns a maximal independent set in $\G$.}
\label{alg:seq-mis}
\end{algorithm}\vspace{1pt}%
\begin{algorithm}[H]
\begin{algorithmic}[1]
	\Function{MIS}{$\G = (V, E)$, $\rank$}
	\State\algorithmicif\ $\lvert V \rvert = 0$ \algorithmicthen\ \Return $\emptyset$\label{line:mis-check}
	\State $S \gets \{v \in V \mid\forall u \in \neigh(v).\ \rank(v) < \rank(u) \}$\label{line:mis-hop}
	\State $R \gets V \setminus \bigcup_{v \in S} \neigh[v]$\label{line:mis-restrict}
	\State\Return $S \cup {}$\Call{MIS}{$\G[R]$, $\rank$}\label{line:mis-rec}
	\EndFunction 
\end{algorithmic}
\caption{Parallel Greedy \textsc{MIS} algorithm, from~\citet{blelloch_greedy_2012}. Given a graph $\G$ and a node ranking $\rank$, returns a maximal independent set in $\G$.}
\label{alg:mis}
\end{algorithm}
\end{table}

\section{Deferred Proofs}

\subsection{Equivalence with downsampling on regular data}

Given the tensors $\tens{V} \in \real^{d_1 \times \dots \times d_n \times f}$, $\tens{K} \in \real^{c_1 \times \dots \times c_n \times f \times g}$, and $\tens{Z}\in \real^{(d_1 - c_1 + 1) \times \dots \times (d_n - c_n + 1) \times g}$, a $n$-dimensional multi-channel convolution~\cite{goodfellow_deep_2016}, can be defined as
\begin{align*}
    \tens{Z}_{i_1, \dots, i_n, k} &= \big[\!\operatorname{conv}(\tens{V}, \tens{K})\big]_{i_1, \dots, i_n, k} \\
    &=\sum_{j_1, \dots, j_n, h} \tens{V}_{i_1 + j_1, \dots, i_n + j_n, h} \cdot \tens{K}_{j_1, \dots, j_n, h, k},
\end{align*}
where $\tens{V}$, $\tens{Z}$, and $\tens{K}$, represent, respectively, the \emph{input}, the \emph{output}, and the \emph{kernel} tensors. (All tensor indices start from~0.) Typically, a zero-padding is also applied to the input tensor, in a way to produce an output tensor with the same shape of the input one~\cite[also known as ``same'' convolution,][]{goodfellow_deep_2016}.

A \emph{strided} convolution, with stride $s \ge 1$, applies the kernel every $s$-th entry of the input tensor, along every (non-channel) dimension, skipping the other entries, i.e., 
\begin{align*}
    \tens{Z}_{i_1, \dots, i_n, k} &= \big[\!\operatorname{conv}(\tens{V}, \tens{K}, s)\big]_{i_1, \dots, i_n, k} \\
    &= \sum_{j_1, \dots, j_n, h} \tens{V}_{s\cdot i_1 + j_1, \dots, s\cdot i_n + j_n, h} \cdot \tens{K}_{j_1, \dots, j_n, h, k}.
\end{align*}
Notice that the strided convolution can be obtained also by \emph{downsampling} the standard one, as
\begin{align*}
\big[\!\operatorname{conv}(\tens{V}, \tens{K}, s)\big]_{i_1, \dots, i_n, k} = \big[\!\operatorname{conv}(\tens{V}, \tens{K})\big]_{s\cdot i_1, \dots, s\cdot i_n, k}.
\end{align*}

On the other hand, a \emph{pooling} layer, with pooling size $p \ge 1$, reduces the input tensor by aggregating every $p$-sided sub-tensor in $\tens{V}$ taken at regular intervals of length $p$, that is,
\begin{align*}
    \tens{Z}_{i_1, \dots, i_n, k} &= \big[\!\operatorname{pool}(\tens{V}, p)\big]_{i_1, \dots, i_n, k} \\
    &= \bigoplus_{j_1, \dots, j_n \in \{0, \dots, p-1\}} \tens{V}_{p\cdot i_1 + j_1, \dots, p\cdot i_n + j_n, k},
\end{align*}
where $\oplus$ is a permutation invariant aggregation function (e.g., \emph{max} or \emph{mean}).

We can model the adjacencies of a tensor's entries by means of a \emph{diagonal grid}, that we define as follows.

\begin{definition}[Diagonal grid]\label{def:diagonal-grid}
Let $d_1, \dots, d_n \in \nat$. A \emph{diagonal grid} $\fullgrid(d_1, \dots, d_n) = (V, E)$ is the graph with vertices
\[
  V = \{0,\dots, d_1 -1\} \times \dots \times \{0,\dots, d_n - 1\}\ (\subset \nat_0^n)
\]
that has an edge joining every two vectors at unit Chebyshev distance, i.e.,
$
    E = \{(\vec{i},\, \vec{j}) \in V^2 \mid \lVert \vec{i} - \vec{j}\rVert_\infty = 1\}.
$
\end{definition}
\begin{remark}\label{rmk:grid-dist}
In a diagonal grid $\fullgrid(d_1, \dots, d_n)$, the distance between two elements $\vec{i}, \vec{j}$ is given by their Chebyshev distance, i.e., $\length(\vec{i}, \vec{j}) = \lVert \vec{i} - \vec{j}\rVert_\infty = \max_h \lvert i_h - j_h\rvert.$
\end{remark}

Notice that the nodes of a diagonal grid $\G = (V, E) = \fullgrid(d_1, \dots, d_n)$ can be used to index the first $n$ dimensions of a tensor $\tens{V} \in \real^{d_1 \times \dots \times d_n \times f}$. As a consequence, this tensor can be adopted as a \emph{feature tensor} of the graph $\G$, and, for any node $\vec{i}\in V$, we can retrieve its feature vector $\vec{x} \in \real^f$ with 
$
\vec{x}_k = \tens{V}_{i_1,\dots,i_n,k}.
$

In \cref{thm:stride} we will prove that, for any $k\ge0$, if $\rank$ is the ranking of the nodes in $V$ in \emph{lexicographic order} (see \cref{def:lex} below), we have that
\begin{enumerate}
    \item $k\text{-MIS}(\G, V, \rank)$ selects the same entries of a strided convolution with $s=k+1$, that is, it will select all the entries of $\tens{V}$ having indices that are multiple of $k+1$ in all the first $n$ dimensions (\cref{thm:stride}.\ref{thm:stride-1}).
    \item $\mathcal{P} = \textsc{Cluster}(\G, k, \rank)$ partitions the tensor as a pooling layer with $p = k+1$, that is, the entries indexed by the vectors in every partition $P\in \mathcal{P}$ will form a $(k+1)$-sided sub-tensor of $\tens{V}$ (\cref{thm:stride}.\ref{thm:stride-2a}).
    \item Finally, the reduced graph obtained by contracting $\mathcal{P}$ (that is, the \emph{quotient graph} $\G/\mathcal{P}$) is isomorphic to the diagonal grid having the same shape (in the first $n$-dimensions) of the output tensor $\tens{Z} = \operatorname{pool}(\tens{V}, k+1)$. Moreover, every partition in $\mathcal{P}$ can be mapped to the aggregated entry in $\tens{Z}$ by a properly defined isomorphism  (\cref{thm:stride}.\ref{thm:stride-2b}).
\end{enumerate}

\begin{definition}[Lexicographic order]\label{def:lex}
Given a finite set of vectors $X \subset \real^n$, the \emph{lexicographic order} on $X \times X$ is defined as
\[
 \vec{x} \prec_{\text{lex}} \vec{y} \iff \bigvee_{i} \Big( (x_i < y_i) \wedge \bigwedge_{j < i} (x_j = y_j) \Big).
\]
\end{definition}

\begin{proposition}\label{thm:stride}
Let $\G = (V, E) = \fullgrid(d_1, \dots, d_n)$ be a diagonal grid, and $\rank$ the ranking of its vertices in lexicographic order.
Then, the following propositions hold for any $k \ge 0\!:$
\begin{enumerate}
    \item\label{thm:stride-1} $S = k\textnormal{-MIS}(\G, V, \rank) = \{ \vec{i} \in V \!\mid \forall h.\, i_h \bmod k + 1 = 0\};$ 
    
    \item\label{thm:stride-2} Let $\mathcal{P} = \textsc{Cluster}(\G, k, \rank)$. Then:
    \begin{enumerate}
        \item\label{thm:stride-2a} $\mathcal{P} = \{\{\vec{i} + \vec{h} \in V \mid \vec{h}\in \{0,\dots,k\}^n \}\}_{\vec{i} \in S};$
        
        \item\label{thm:stride-2b} $\G/\mathcal{P} \cong \fullgrid\!\big(\big\lfloor\frac{d_1}{k+1}\big\rfloor, \dots, \big\lfloor\frac{d_n}{k+1}\big\rfloor\big)$ with isomorphism 
        \[
        \phi(P) = \tfrac{1}{k+1}\cdot \vec{i} \quad\text{such that}\quad \{\vec{i}\} = P \cap S.
        \]
        %
    \end{enumerate}
\end{enumerate}
\end{proposition}
\begin{proof}
\begin{enumerate}
    \item We can prove the first point by well-founded induction on the sequence of nodes ordered by their ranking $\rank$. Specifically, we prove the following proposition for every $u \in V$:
    \begin{align*}
    P(u) &\equiv S_u = \{ \vec{i} \in V_u \mid \forall h.\ {i}_h \bmod k + 1 = 0\},
    \end{align*}
    where $V_u = \{v \in V \mid \rank(v) \le \rank(u)\}$ and $S_u = k\textnormal{-MIS}(\G, V_u, \rank)$. Notice that $V_u$ represents also the set of the first $\lvert V_u\rvert$ vertices processed by \cref{alg:k-mis}, in its equivalent sequential implementation, and hence $S_u \subseteq S$ for every $u\in V$.
    \begin{description}
    \item[Base case.] The minimal node with respect to the lexical ordering is $\vec{0} \in V$. $P(u)$ is satisfied, since 
    $u =\vec{0}$ is the first and only element in $V_u$, which is then selected and returned as singleton $k$-independent set by \cref{alg:k-mis}.
    
    \item[Induction.] For any $u \in V,$ we assume by induction that $P(w)$ holds for any other $w \in V_u \setminus \{u\}$. Let $v$ be the maximal element in this set, i.e., 
    \[
    v = \operatorname{argmax}_{w \in V_u\setminus \{u\}} \rank(w). 
    \]
    %
    Since \cref{alg:k-mis} selects the vertices in lexicographic order, $u$ will be the last one extracted from $V_u$. Hence, the returned $k$-independent set will be either
    \begin{align*}
        k\textnormal{-MIS}(\G, V_u, \rank) = \begin{cases}
        S_v & \text{if } \length(u, S_v) \le k,\\
        S_v \cup \{u\} & \text{otherwise.}
        \end{cases}
    \end{align*}
    In the first case, $P(u)$ is satisfied, since any node in $S$
    %
    is at least $k+1$ hops from any other node in $S_v\, (\subseteq S)$. 
    In the second case, assume by contraposition that $u =\vec{j}\not\in S$. Then, there exists at least an index $h$ such that $j_h \bmod k+1 \neq 0$. Let $\vec{j}' \in S$ be defined as
    \begin{align}\label{eq:nearest-centroid}
    \forall i.\ j'_i = {j}_i - (j_i \bmod k + 1) \le j_k.
    \end{align}
    This index precedes lexicographically $\vec{j}$ and must belong to a node in $S_v$ by inductive assumption. This is impossible since $\length(\vec{j}, \vec{j}') = \lVert \vec{j} - \vec{j}'\rVert_\infty\le k$ violates the $k$-independence condition, hence $\vec{j}$ must be also in $S$, and $P(u)$ is satisfied. 
    \end{description}
    
    \item[\ref{thm:stride-2a}.] \cref{alg:cluster} associates every node in $V$ to the $k$-hop neighbor in $S = k\textnormal{-MIS}(\G, V, \rank)$ with lowest rank. We know by {\cref{thm:stride}.\ref{thm:stride-1}} that $S = \{ \vec{i} \in V \mid \forall h.\ i_h \bmod k + 1 = 0\}$ and, by \cref{rmk:grid-dist}, that the (inclusive) $k$-hop neighborhood of a node $\vec{i}$ is given by
    \begin{align*}
    \neigh_{\G}[\vec{i}] &= \{\vec{j} \mid \lVert \vec{i} - \vec{j}\rVert_\infty \le k \} \\
    &= \{\vec{i} + \vec{h} \mid \vec{h} \in \mathbb{Z}^n\ \wedge\ \lVert\vec{h}\rVert_\infty \le k \}.
    \end{align*}
    For every node in $V$, the $k$-hop neighbor in $S$ with lowest rank will always be the one having offset $\vec{h}$ with all non-positive entries (there is always one, and can be retrieved as in \cref{eq:nearest-centroid}). Dually, every $\vec{i} \in S$ will form a cluster with all of its $k$-hop neighbors having offset $\vec{h} \in \{0,\dots,k\}^n.$
    
    \item[\ref{thm:stride-2b}.] Let $\G / \mathcal{P} = (\mathcal{P}, \mathcal{E})$ and $\fullgrid\!\big(\big\lfloor\frac{d_1}{k+1}\big\rfloor, \dots, \big\lfloor\frac{d_n}{k+1}\big\rfloor\big) = (V', E')$. 
    By \cref{thm:stride}.\ref{thm:stride-1} and \ref{thm:stride}.\ref{thm:stride-2a}, the cluster of any node $\vec{i} \in S$ will be connected to the ones of any other node $\vec{j}\in S$ such that $\lVert \vec{i} - \vec{j} \rVert_\infty = k+1$. These are also the only edges in the quotient graph since, by maximality of the $k$-independent set, there is an edge between two clusters $P, Q \in \mathcal{P}$ if and only if the two vertices $\{\vec{i},\vec{j}\} = S \cap (P \cup Q)$ are at most $2k+1 < 2(k+1)$ apart from each other (otherwise a node between $\vec{i}$ and $\vec{j}$ would belong to a different cluster). The function $\phi: \mathcal{P} \to V'$ is a bijection since 
    \begin{enumerate}
        \item any cluster in $\mathcal{P}$ has one and only one node in $S$ ($\mathcal{P} \leftrightarrow S$), and
        \item by scaling the vectors in $S$ by $k+1$ we obtain exactly the node set $V'$ ($S \leftrightarrow V'$).
    \end{enumerate}
    Hence, we can reach the conclusion as following,
    \begin{align*}
        (P,\, Q) \in \mathcal{E} &\Leftrightarrow \lVert \vec{i} - \vec{j} \rVert_\infty = k + 1\tag{$\vec{i}, \vec{j}\in S\cap(P\cup Q)$}\\
        &\Leftrightarrow \lVert (k+1) \cdot (\phi(P) - \phi(Q))\rVert_\infty = k + 1 \\ 
        &\Leftrightarrow \lVert \phi(P) - \phi(Q)\rVert_\infty = 1 \\
        &\Leftrightarrow (\phi(P),\, \phi(Q)) \in E'.
    \end{align*}
    
\end{enumerate}
\end{proof}

\subsection{Connectivity}\label{sec:connectivity-proofs}


\begin{lemma}\label{lemma:length}
	Let $\G$ be a connected graph and $\H = \reduce\big(\G; k\big)$, with $k \in \nat$. Then, $\forall u, v \in V(\H)$,
	\[
	\length_{\H}(u, v) \le \length_\G(u, v) \le (2k + 1)\length_{\H}(u, v).
	\]

\end{lemma}
\begin{proof} 
The first inequality 
is proven by construction of $H$: for each node $w$ in the shortest path $\P = u{\sim}v \subseteq \G$ (including $u$ and $v$) there exist a node $w' \in V(\H)$ such that $\length_{\G}(w, w') \le k$;
%
let $R$ the set of these nodes. Each pair of adjacent nodes in $\P$ either belong in the same $k$-hop neighborhood for some node $w' \in R$, or belong in two distinct ones $w',w''$ which are made adjacent by the coarsening process. Hence the path induced by $R$ is connected and of size at most $e(\P) = \length_\G(u, v)$, so $\length_\H(u, v) \le |R| - 1 \le \length_{\G}(u, v)$.  


For the second inequality, let $\P' = u{\sim}v$ a shortest path in $\H$, of length $\length_\H(u, v)$; recalling Remark~\ref{rmk:centroid_hops}, any two consecutive nodes $u, w$ on $\P'$ are at (hop) distance at most $2k + 1$ in $\G$, so there is a $u,v$-path in $\G$ of length at most $(2k + 1)\length_\H(u, v)$.
\end{proof}


%
\begin{proof}[Proof of \cref{thm:length}]
The first inequality holds by the same arguments used for the fist inequality of \cref{lemma:length}. The second inequality follows from \cref{lemma:length} and triangle inequality. Namely,
\begin{align*}
    \length_\G(u, v) \le \length_\G(u, \repr(u)) + \underbrace{\length_{\G}(\repr(u), \repr(v))}_{\hspace{-1em}\le (2k + 1)\length_{\H}(\repr(u), \repr(v))\hspace{-1em}} + \length_\G(\repr(v), v).
\end{align*}
Finally, $\forall x\in V,\ \length_\G(\repr(x), x)\le k$ by construction of $H$.
\end{proof}

\begin{proof}[Proof of \cref{th:cc}]
If $\G$ is connected, then also $\H$ is connected, by \cref{lemma:length}. Otherwise, let $C \subseteq V(\G)$ the nodes of a connected component of $\G$. Since $V(\H)$ is a \kmis of $\G$, $C \cap V(H)$ is a \kmis of $\G[C]$. Applying \cref{alg:cluster} to $\G[C]$ (with the same ordering used on $\G$) will produce the reduced graph $\H' = \H[C \cap V(\H)]$, which by \cref{lemma:length} is connected. Finally, \cref{alg:cluster} joins two nodes in $V(\H)$ only if there exists and edge in $\G$ intersecting their $k$-hop neighborhood, hence \cref{alg:cluster} does not connect different components of $\G$.
\end{proof}

\subsection{Lower bounds}\label{sec:bound-proofs}


\begin{definition}[Restricted Neighborhood]\label{def:rneigh}
Let $\G = (V, E)$ be a graph, $\rank$ be a ranking of its nodes, and $k \in \nat$ an integer. Compute $S = k\textnormal{-MIS}(\G, V, \rank)$ as in \cref{alg:k-mis}, and let $V_0 \supseteq \dots \supseteq V_r$ and $S_0 \supseteq \dots \supseteq S_r$ the filtration of nodes produced respectively by the inputs and the outputs of its $r$ recursive calls, with $V_0 = V$, $S_0 = S$, and $V_r = S_r = \emptyset$. We define the \emph{restricted $k$-hop neighborhood} of a node $v \in S$, denoted $\rneigh_k(v)$ (resp.\ $\widehat{\neigh}_k[v]$ if inclusive), as
\begin{align*}
    \rneigh_k(v) = \neigh_k(v) \cap V_{i^\ast} \quad\text{ with }\quad  i^\ast = \max \{ i \mid v \in S_i\}.
\end{align*}
\end{definition}

\begin{remark}\label{th:rneigh-partition}
Let $S = k\textnormal{-MIS}(\G, V, \rank)$. Then, $\mathcal{P} = \{\rneigh_k[v]\}_{v\in S}$ forms a partition of $V$.
\end{remark}%

\begin{remark}\label{th:deg-plus-one}
Let $\G = (V, E)$ a graph, $\adj \in \{0, 1 \}^{n\times n}$ its (unweighted) adjacency matrix, and $\vec{x} \in \real_+^n$ a vector of non-negative values. Then, for any $v \in V$ and $k \in \nat$, %
\[
\sum_{u \in \neigh_{k}[v]} \vec{x}_u \le [(\adj + \I)^k\vec{x}]_v.
\]  
\end{remark}%

\begin{proof}[Proof of \cref{th:bound-k-degree}]
\begin{align*}
    \textstyle\sum_{v \in S} \vec{x}_v &= \textstyle\sum_{v \in S} w(v)\cdot [(\adj + \I)^k\vec{1}]_v \\
    &\ge \textstyle\sum_{v \in S} w(v) \cdot\big\lvert \neigh_k[v]\big\rvert\tag{\cref{th:deg-plus-one}} \\
    &\ge \textstyle\sum_{v \in S} w(v) \cdot\big\lvert \rneigh_k[v]\big\rvert\tag{$\rneigh_k[v] \subseteq \neigh_k[v]$} \\
    &\ge \textstyle\sum_{v \in S} \textstyle\sum_{u \in \rneigh_k[v]} w(u)\tag{$\forall u\in\rneigh_k[v].\ w(v) \ge w(u)$} \\
    &= \textstyle\sum_{v\in V} w(v).\tag{\cref{th:rneigh-partition}}
\end{align*}
\end{proof}

\begin{proof}[Proof of \cref{th:bound-k-weights}]
\begin{align*}
    \textstyle\sum_{v \in S} \vec{x}_v &= \textstyle\sum_{v \in S} w(v)\cdot [(\adj + \I)^k\vec{x}]_v \\
    &\ge \textstyle\sum_{v \in S} w(v) \cdot \sum_{u \in \neigh_k[v]}\vec{x}_u\tag{\cref{th:deg-plus-one}} \\
    &\ge \textstyle\sum_{v \in S} w(v) \cdot \textstyle\sum_{u \in \rneigh_k[v]}\vec{x}_u\tag{$\rneigh_k[v] \subseteq \neigh_k[v]$} \\
    &\ge \textstyle\sum_{v \in S} \textstyle\sum_{u \in \rneigh_k[v]} w(u) \cdot \vec{x}_u\tag{$\forall u\in\rneigh_k[v].\ w(v) \ge w(u)$} \\
    &= \textstyle\sum_{v\in V} w(v) \cdot \vec{x}_v.\tag{\cref{th:rneigh-partition}}
\end{align*}
\end{proof}

\begin{remark}\label{th:associativity}
Let $\mat{A} \in \real^{n\times n}$ be a symmetric matrix and $\vec{x} \in \real^n$ be a vector. Then
\[
\sum_{i = 1}^n \sum_{j = 1}^n \mat{A}_{ij}\cdot\vec{x}_j = \sum_{i = 1}^n \vec{x}_i \cdot \sum_{j = 1}^n \mat{A}_{ij}.
\]
\end{remark}

\begin{proposition}[\citet{kako_approximation_2009}]\label{th:kako-prop}
Assume that $a_i > 0$ and $b_i > 0$ for $1 \le i \le n$. Then
\[
\sum_{i = 1}^n\frac{{b_i}^2}{a_i} \ge \frac{\big(\sum_{i = 1}^n b_i\big)^2}{\sum_{i = 1}^n a_i}.
\]
\end{proposition}

\begin{proof}[Proof of \cref{th:ratio}]
Let $\Delta_k = \max_{v \in V}\ [(\adj + \I)^k\vec{1}]_v$.
When using a ranking induced by \cref{eq:rank-k-degree} we have that
\begin{align*}
    \textstyle\sum_{v\in S}\vec{x}_v &\ge \textstyle\sum_{v \in V} w(v)\tag{\cref{th:bound-k-degree}}\\
    &\ge  \frac{\sum_{v \in V} \vec{x}_v}{\Delta_k}\tag{$\forall v. \in V.\ \Delta_k \ge [(\adj + \I)^k\vec{1}]_v$}\\
    &\ge \frac{\alpha(G^k)}{\Delta_k}.\tag{$\sum_{v \in V} \vec{x}_v \ge \alpha(\G^k)$}
\end{align*}
When using the rank induced by \cref{eq:rank-k-weights}, instead, we have
\begin{align*}
    \sum_{v\in S}\vec{x}_v &\ge\sum_{v \in V} \frac{{\vec{x}_v}^2}{[(\adj + \I)^k\vec{x}]_v}\tag{\cref{th:bound-k-degree}}\\
    &\ge \frac{\big(\sum_{v \in V} \vec{x}_v\big)^2}{\sum_{v \in V} [(\adj + \I)^k\vec{x}]_v}\tag{\cref{th:kako-prop}}\\
    &= \frac{\big(\sum_{v \in V} \vec{x}_v\big)^2}{\sum_{v \in V} \vec{x}_v \cdot [(\adj + \I)^k\vec{1}]_v}\tag{\cref{th:associativity}}\\
    &\ge \frac{\big(\sum_{v \in V} \vec{x}_v\big)^2}{\Delta_k\sum_{v \in V} \vec{x}_v }\tag{$\forall v. \in V.\ \Delta_k \ge [(\adj + \I)^k\vec{1}]_v$}\\
    &\ge \frac{\alpha(G^k)}{\Delta_k}.\tag{$\sum_{v \in V} \vec{x}_v \ge \alpha(\G^k)$}
\end{align*}

\end{proof}

\section{Experimental Setting}

\subsection{Description of the datasets}\label{subsec:graph-descr}

We run benchmark experiments on real-world undirected graphs from different domains, namely
\begin{itemize}
    \item \emph{Orkut}, \emph{LiveJournal}, \emph{Youtube}, and \emph{Brightkite}, are social networks from the SNAP dataset~\cite{leskovec_snap_2014}, where every node represents a {user} and every edge a {friendship} relation. Like most social networks, those networks have an small effective diameter, that in this case amounts at most to $6.5$.
    \item \emph{Skitter}, which is an Internet topology graph built from traceroutes, where every autonomous system (AS) is represented as a node which is connected by an edge to other ASs if there there has been reported an exchange of information between two of them. This graph was also retrieved from the SNAP dataset~\cite{leskovec_snap_2014}. Its diameter is $6$.
    \item \emph{Enron} is an email communication network from SNAP~\cite{leskovec_snap_2014}, where two nodes are email addresses and there is an edge between two of them if they exchanged at least an email (sent or received). It has an effective diameter of $4.8$.
    \item \emph{DBLP} and \emph{AstroPh} are two co-authorship networks, respectively from the 10th DIMACS Challenge~\cite{bader_10th_2011} and from the SNAP~\cite{leskovec_snap_2014} dataset. In \emph{DBLP}, every node in the networks represents a paper and there is an edge if two papers share at least an author, while in \emph{AstroPh} every node represents an author and there is an edge between two of them if they co-authored a paper. Their effective diameters are $6.8$, and $4.8$, respectively.
    \item \emph{Europe} and \emph{Luxembourg} are two road networks from the 10th DIMACS Challenge dataset~\cite{bader_10th_2011}. In these networks every edge represents a road (of some kind) and a node a crossing. These are the only weighted graphs, where every weight represents the Euclidean distance between the coordinates of the two endpoints. Differently from the previous networks, street maps are (almost) planar graphs and, as such, can be expected to have a high diameter. 
\end{itemize}
All the graphs were retrieved from the University of Florida Sparse Matrix Collection~\cite{davis_university_2011}.

For the classification tasks, we used the following benchmark datasets:
\begin{itemize}
    \item \emph{DD}~\cite{dobson_distinguishing_2003}, a dataset of graphs representing protein structures, in which the nodes are (labeled) amino acids and two nodes are connected by an edge if they are less than 6 Angstroms apart. The task consists in classifying enzymes and non-enzymes.
    \item \emph{REDDIT}~\cite[\emph{-BINARY}, \emph{-MULTI-5K}, and \emph{-MULTI-12K},][]{yanardag_deep_2015}, are social networks where there is an edge between two users if there was reported an interaction between them (in the form of comments in a discussion thread). The task consists in classifying different kinds of communities.
    \item \emph{GITHUB-STARGAZERS}~\cite{rozemberczki_karate_2020}, a social network of developers, where every edge is a ``following'' relation. The task is to decide if a community belongs to web or machine-learning developers.
\end{itemize}
Since REDDIT and GITHUB datasets have no node labels, we set them to a constant value (fixed to 1) for every node.
All the benchmark datasets were retrieved from the \emph{TUDataset} collection~\cite{morris_tudataset_2020}. Further information regarding the graphs and datasets are summarized in \cref{tab:graph-info}. 
\begin{table}[htb]
    \centering
    \resizebox{\columnwidth}{!}{%
    \footnotesize
\begin{tabular}{ll@{\ \ }rrrr}
\toprule
Graph & Type &        $n$ &        $m$  & $\degree$ & $\length_{90}$\\
\midrule
Orkut         & Social & 3072441 & 117185083         & $76.3$ & $4.8$ \\
{LiveJournal} & Social      & 3997962 &   69362378 & $34.7$ & $6.5$ \\
{Youtube}     & Social      & 1134890 	 &    5975248 & $10.5$ & $6.5$ \\
{Brightkite}  & Social    & 58228  &   214078   &  $7.4$ & $6.0$ \\
{Skitter}     & Web    & 1696415 &   22190596 & $26.2$ & $6.0$ \\
{Enron}       & Email    & 36692  &   367662   & $20.0$ & $4.8$ \\
{AstroPh}     & Auth.    & 18772  &   396160   & $42.2$ & $4.8$ \\
{DBLP}        & Auth.    & 540486 &   30491458 & $112.8$ & $6.8$ \\
{Europe}      & Road        & 50912018 &  108109320 &  $4.2$ & $>$$10^3$ \\
{Luxembourg}  & Road        & 114599   &  239332    &  $4.2$ & $\approx$$10^3$ \\ 
\midrule\midrule
    Dataset & Type & $n$ (avg.) &  $m$ (avg.) & Size & Class  \\\midrule
    DD &  Protein &  284.32 & 715.66 & 1178 & 2 \\
    REDDIT-B &  Social &  429.63  & 497.75 & 2000 & 2 \\
    REDDIT-5K &  Social &  508.52  & 594.87 & 4999 & 5 \\
    REDDIT-12K &  Social &  391.41  & 456.89 & 11929 & 11 \\
    GITHUB &  Social &  113.79  & 234.64 & 12725 & 2 \\
\bottomrule
    \end{tabular}}
    \caption{Benchmark graphs and dataset information}
    \label{tab:graph-info}
\end{table}

\subsection{Ablation studies}
\begin{table*}[h]
    \centering
    \small
\begin{tabular}{llr@{$\,\pm\,$}lr@{$\,\pm\,$}lr@{$\,\pm\,$}lr@{$\,\pm\,$}lr@{$\,\pm\,$}l} 
\toprule
Model & Sampling & \multicolumn{2}{c}{DD} & \multicolumn{2}{c}{REDDIT-B} & \multicolumn{2}{c}{REDDIT-5K} & \multicolumn{2}{c}{REDDIT-12K} & \multicolumn{2}{c}{GITHUB} \\
\midrule
TopKPool     & {Top-$k$} & {\bfseries 74.92} & {\bfseries 2.03} &         81.10 & 3.82 &           45.28 & 3.88 &            38.55 & 2.35 &  65.93 & 0.45 \\
             & {$k$-MIS} & 74.75 & 1.64 &         {\bfseries 86.85} & {\bfseries 0.41} &           {\bfseries 52.80} & {\bfseries 0.62} &            {\bfseries 46.42} & {\bfseries 0.43} &  {\bfseries 67.32} & {\bfseries 0.81} \\\midrule
SAGPool      & {Top-$k$} & 73.26 & 2.26 &         84.90 & 3.94 &           46.29 & 5.61 &            42.30 & 3.70 &  64.29 & 5.70 \\
             & {$k$-MIS} & {\bfseries 75.97} & {\bfseries 1.59} &         {\bfseries 88.87} & {\bfseries 2.18} &           {\bfseries 52.77} & {\bfseries 1.82} &            {\bfseries 47.69} & {\bfseries 0.68} &  {\bfseries 68.75} & {\bfseries 0.39} \\\midrule
ASAPool      & {Top-$k$} & 73.73 & 2.18 &         {\bfseries 78.37} & {\bfseries 5.22} &           39.53 & 7.76 &            {\bfseries 39.14} & {\bfseries 3.58} &  {\bfseries 66.98} & {\bfseries 0.96} \\
             & {$k$-MIS} & {\bfseries 74.75} & {\bfseries 1.20} &         76.36 & 6.25 &           {\bfseries 48.20} & {\bfseries 1.57} &            34.23 & 6.29 &  66.30 & 0.33 \\\midrule
PANPool      & {Top-$k$} & 73.26 & 1.94 &         77.44 & 4.95 &           46.04 & 3.78 &            {40.97} & {3.02} &  62.48 & 2.84 \\
             & {$k$-MIS} & {\bfseries 76.23} & {\bfseries 1.97} &         {\bfseries 85.85} & {\bfseries 1.06} &           {\bfseries 49.63} & {\bfseries 1.47} &            {\bfseries 45.10} & {\bfseries 0.84}
             &  {\bfseries 64.82} & {\bfseries 1.41} \\
\bottomrule
\end{tabular}
\caption{Classification accuracy on selected benchmark datasets \emph{(mean{\footnotesize $\bm{\,\pm\,}$}std)}.}
\label{tab:ablation}
\end{table*}
The \kmis selection (\cref{alg:k-mis}) and reduction (\cref{alg:cluster}) can also be used as a \emph{plug-in replacement} of the more trivial \topk selection, where, in the first case, $k$ regulates the reduction of the graph as discussed in \cref{sect:kmis}, while in the latter it specifies the number of nodes to be selected from the graphs (usually defined in terms of a fraction of the size of the graphs). In \cref{tab:ablation} we show how changing the sampling mechanism from \topk to \kmis in pooling methods from the literature affects their performance on the selected benchmark datasets. Apart from \model{TopKPool}, where by sampling the \topk nodes it achieves better results only on the DD dataset (the smallest one), we can see instead that substituting the \topk with \kmis selections boost the performance of the consdidered pooling methods. The only methods in which the different sampling seem to provide little to no benefit is \asapool. We argue that this is caused by the more complex (and more expensive) reductions adopted by this method: while \topk methods, once selceted the $S \subseteq V$ nodes from the input graph (with $\lvert S \rvert = k = \lceil rn\rceil$ for some ratio $r \in(0, 1)$), they reduce the adjacency matrix of the graph $\adj\in \real^{n\times n}$ by selecting its rows and column $\adj_\text{out} = \adj_{S, S} \in \real^{k\times k}$, \asapool reduces the input matrix $\adj$ by selecting \emph{also the neighborhoods} of the $S$ selected nodes, defining $\Smat = \adj_S$, which itself could require $O(rn^2)$ space, and then reducing the adjacency matrix by following \citet{ying_hierarchical_2018}, thus computing $\adj_{\text{out}} = \Smat\adj\Smat^{\tr}$. When $\adj$ is symmetric, this is equivalent as computing $\adj_{\text{out}}=\adj_{S,S}^3 \in \real^{k\times k}$, that is, selecting the $S$ nodes on the \emph{third power} of the input graph, which can become dense at will (or even full). As an example, if \asapool selects $k = \lceil rn\rceil$ peripheral nodes of a star graph of $n + 1$ nodes and $n$ edges, the resulting reduced graph becomes complete (i.e., a graph with $k^2 \ll n$ edges). This property may facilitate the exchange of information between nodes in the reduced graph, but will hinder the scalability of the method and, hence, \asapool should not be even be considered as a ``sparse'' pooling method, differently from the other \topk ones or \kmis.

\subsection{Other benchmarks}\label{subsec:exp-time}
\begin{figure*}[th]
    \centering
    \hspace*{-1.5ex}
    \resizebox{1.03\linewidth}{!}{%
    \input{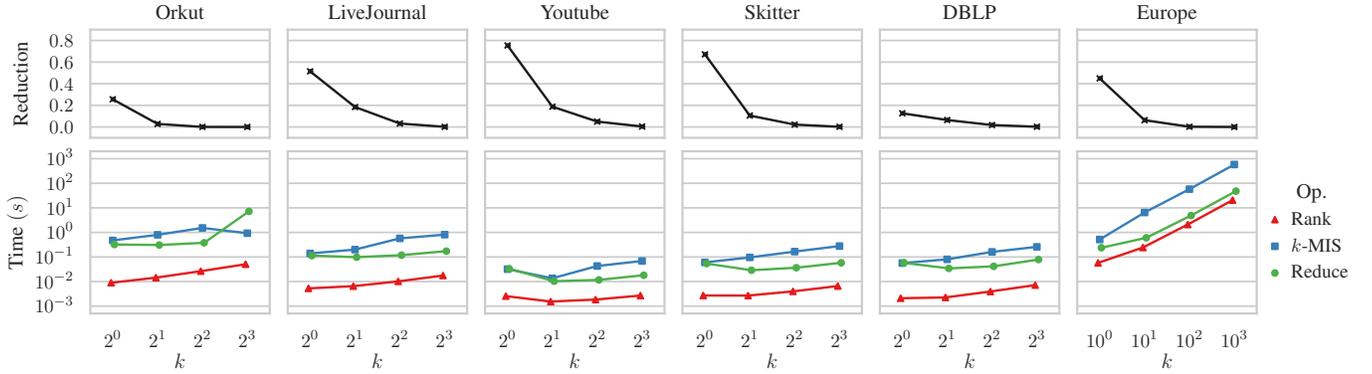}}
    \caption{Reduction ratio (\emph{top}) and running time (\emph{bottom}, in log scale) of our approach for varying $k$ (also in log scale).}
    \label{fig:time}
\end{figure*}
\begin{table}[th]
    \centering
    \resizebox{\linewidth}{!}{%
    \small
\begin{tabular}{llrrrr}
\toprule
 & & \multicolumn{2}{c}{Ranking with \cref{eq:rank-k-degree}} &  \multicolumn{2}{c}{  Ranking with \cref{eq:rank-k-weights}}  \\\cmidrule(lr){3-4}\cmidrule(lr){5-6}
Graph & $k$ & Greedy &       Ours   &   Greedy         &       Ours    \\
\midrule
AstroPh & 1 &   392347.4 &    392355.1 &   392533.8 &   392528.4 \\
           & 2 &   149701.3 &    148257.2 &   149625.2 &   147953.5 \\
           & 3 &    72923.3 &     71929.1 &    72857.1 &    71847.4 \\
\midrule
Enron & 1 &  1220157.4 &   1220194.0 &  1219789.6 &  1219789.3 \\
           & 2 &   217197.2 &    215428.5 &   217131.7 &   215016.6 \\
           & 3 &   148230.3 &    147460.3 &   148175.6 &   147392.6 \\
\midrule
Brightkite & 1 &  1931765.8 &   1931884.3 &  1932375.7 &  1932378.2 \\
           & 2 &   824662.6 &    821815.2 &   824654.3 &   820564.1 \\
           & 3 &   456965.4 &    457761.3 &   456838.0 &   457368.3 \\
\midrule
Luxembourg & 1 &  3286607.7 &   3286587.3 &  3309810.9 &  3309836.2 \\
           & 2 &  2267010.7 &   2265353.6 &  2290198.3 &  2284708.0 \\
           & 3 &  1721185.9 &   1717789.7 &  1740735.1 &  1725635.3 \\
           & 4 &  1379199.0 &   1373352.9 &  1393949.6 &  1371500.4 \\
           & 5 &  1147810.9 &   1141633.0 &  1159119.1 &  1133255.1 \\
           & 6 &   978606.1 &    970727.6 &   988133.8 &   959172.7 \\
           & 7 &   851061.6 &    840502.3 &   858769.6 &   830110.7 \\
           & 8 &   751680.4 &    739632.1 &   757523.8 &   728799.9 \\
\bottomrule
\end{tabular}}
\caption{Weight comparison of ${k}$-MISs obtained with our relaxation and the classical greedy algorithm.} 
\label{tab:mwis-results}
\end{table}
\Cref{fig:time} reports the average reduction ratio (\emph{top row}) and average running time (\emph{bottom row}, in \emph{user} seconds) of ten runs of our method on selected benchmark graphs, using different values of $k$ and a ranking induced by \cref{eq:rank-k-weights} with constant weights. Times refer to the ones needed respectively to compute the ranking (as described in \cref{sect:theo}), the \kmis (\cref{alg:k-mis}), and then reducing the graph (\cref{alg:cluster} and edge reduction, as described in \cref{sect:kmis}). We can clearly observe the linearity of the time complexity of our algorithm, since the execution time increases  linearly with $k$. We can also see that, for the graphs with small diameter (see \cref{tab:graph-info}), the running time decreases once $k$ approaches their effective diameter, since most of the nodes will be assigned to the same centroid during the first recursive call of \cref{alg:k-mis}, thus decreasing the expected depth of the algorithm and also its execution time. 

\cref{tab:mwis-results} reports the average total weight obtained by computing the greedy sequential MIS algorithm on $\G^k$ \emph{(Greedy)}, compared to \cref{alg:k-mis} on $\G$ \emph{(Ours)}. For both algorithms we used \cref{eq:rank-k-degree,eq:rank-k-weights}, fixing $k=1$ for the greedy one (thus applying the original rules of \citet{sakai_note_2003} on $\G^k$). Results are averaged on ten runs with node weights extracted uniformly at random in the interval $[1, 100]$. The rationale behind these results is two-fold.
First, differently from \cref{alg:k-mis}, in sequential greedy algorithms \cite[see, e.g.,][]{sakai_note_2003,kako_approximation_2009} those nodes maximizing \cref{eq:rank-k-degree} or (\ref{eq:rank-k-weights}) are iteratively added to the independent set and their neighbors are removed from the graph. \cref{eq:rank-k-degree,eq:rank-k-weights} are computed on the remaining subgraph, possibly producing a different value with respect to the one computed in the previous step. Nonetheless, in \cref{tab:mwis-results} we can see that, for $k=1$, the difference in the results between computing \cref{eq:rank-k-degree,eq:rank-k-weights} at every step \emph{(Greedy)} or just once \emph{(Ours)} is minimal (we actually obtain a better approximation on certain configurations).
Secondly, even if  \cref{eq:rank-k-degree,eq:rank-k-weights} for $k > 1$ are a loose overestimation of the same formula computed on $\G^k$, we show that the performance deteriorates very slowly with the increase of $k$, obtaining a total weight similar to the one computed by the greedy algorithm on the (denser) graph $\G^k$ for small values of $k$.

\subsection{Model selection}\label{sec:model-selection}

The model architecture, which is shared among all the trained models, can be summarized as follows:
\begin{small}%
\begin{align*}
\text{GNN}_{h} \to \text{Pool} &\to \text{GNN}_{2h} \to \text{Pool} \to \text{GNN}_{4h} \to \text{Glob} \to \text{MLP}, 
\end{align*}%
\end{small}%
where $h$ is number of output features, $c$ is the number of classes in the dataset, Glob is the global \emph{sum} and \emph{max} aggregation of the remaining node features concatenated, while MLP is defined as
\begin{small}%
\[
\text{Dropout}_{p=0.3} \to \text{NN}_{2h} \to \text{Dropout}_{p=0.3} \to \text{NN}_{c}.
\]%
\end{small}%
Every GNN/NN is followed by a \emph{ReLU} activation function. The pooling method (Pool) is changed for every method (and removed for the baseline), while the GNN layer is chosen among a set of layers during the model selection phase. The possible GNN layers are \textsc{GCN}~\cite{kipf_semi-supervised_2017}, \textsc{GATv2}~\cite{brody_how_2021}, and \textsc{GIN}~\cite{xu_how_2019}, unless the model is \textsc{PANPool}, which requires a specific GNN layer \cite[\textsc{PANConv},][]{ma_path_2020}. The best model for each poling layer was chosen using a grid search, with hyper-parameter space defined as in \cref{tab:hyper-params}, where $\eta$ refers to the learning rate, $b$ to the batch size, $r$ to the reduction ratio, $L$ to the maximal path length of \textsc{PANConv}, and $\sigma$ to the scoring activation function of \textsc{EdgePool}, as proposed by its authors~\cite{diehl_towards_2019}. To increase the parallelization of the grid search, every configuration run had a cap of 8GB of GPU memory to compute the training. Runs that could not fit into this limit were discarded from the model selection process.
\begin{table}[htb]
    \centering
    \resizebox{\linewidth}{!}{%
    \small
    \begin{tabular}{lll}
    \toprule
    Models & Param. & Values \\\midrule
    All & $\eta$ & 0.001, 0.0001 \\
        & $h$ & 32, 64, 128 \\
        & $b$ & 32, 64, 128 \\\midrule
    All except \textsc{PANPool} & GNN & \textsc{GCN}, \textsc{GATv2}, \textsc{GIN} \\\midrule
    \textsc{PANPool} & GNN & \textsc{PANConv}\\
     & $L$ & 1, 2, 3 \\\midrule
    \textsc{TopK}-, SAG-, ASAP-, \textsc{PANPool} & $r$ & 0.5, 0.2, 0.1 \\\midrule
    \textsc{EdgePool} & $\sigma$ & \emph{tanh}, \emph{softmax} \\\midrule
    \kmis, BDO & $k$ & 1, 2, 3 \\\bottomrule
    \end{tabular}}
    \caption{Hyper-parameter space}
    \label{tab:hyper-params}
\end{table}

\subsection{Hardware and software}

Our method has been implemented using PyTorch~\cite{paszke_pytorch_2019} and PyTorch Geometric~\cite{fey_fast_2019}, to allow high-level scripting of CUDA code. Every label propagation required by the algorithms has been implemented in form of message-passing (i.e., gather-scatter) to run exclusively on GPU. All the experiments has been executed on a machine running Ubuntu Linux with an AMD EPYC 7742 64-Core processor with 1TB of RAM, and a NVIDIA A100 with 40GB of on-board memory.
}{}
\end{document}